\documentclass[10pt,journal,compsoc]{IEEEtran}
\usepackage[utf8]{inputenc}
\usepackage{amsmath}
\usepackage{hyperref}
\usepackage{tabularx}
\usepackage{graphicx}
\usepackage{subfig}

\usepackage{amssymb}

\usepackage{verbatim}

\usepackage{multirow}

\usepackage{xargs}                      
\usepackage[pdftex,dvipsnames]{xcolor}  
\usepackage[colorinlistoftodos,prependcaption,textsize=tiny]{todonotes}

\ifCLASSOPTIONcompsoc
  \usepackage[nocompress]{cite}
\else
  \usepackage{cite}
\fi

\newcommandx{\unsure}[2][1=]{\todo[linecolor=red,backgroundcolor=red!25,bordercolor=red,#1]{#2}}
\newcommandx{\change}[2][1=]{\todo[linecolor=blue,backgroundcolor=blue!25,bordercolor=blue,#1]{#2}}
\newcommandx{\info}[2][1=]{\todo[linecolor=OliveGreen,backgroundcolor=OliveGreen!25,bordercolor=OliveGreen,#1]{#2}}
\newcommandx{\improvement}[2][1=]{\todo[linecolor=Plum,backgroundcolor=Plum!25,bordercolor=Plum,#1]{#2}}
\newcommandx{\thiswillnotshow}[2][1=]{\todo[disable,#1]{#2}}

\begin{document}
\title{Detect Faces Efficiently: A Survey and Evaluations}

\author{Yuantao~Feng,
        Shiqi~Yu$^*$,
        Hanyang~Peng,
        Yan-Ran~Li,
        Jianguo~Zhang.
\thanks{Manuscript received November 7, 2020; revised February 28, 2021 and June 14, 2021; accepted September 15, 2021. The work was supported in part by the National Natural Science Foundation of China (Grant No. 61976144 and 61806128), the National Key Research and Development Program of China (Grant No. 2020AAA0140002) and the Stable Support Plan Program of Shenzhen Natural Science Fund (Grant No. 20200925155017002).\textit{(Corresponding author: Shiqi Yu, yusq@sustech.edu.cn)}}

\thanks{Yuantao Feng, Shiqi Yu, Hanyang Peng and Jianguo Zhang are with the Department of Computer Science and Engineering, Southern University of Science  and Technology, China.}

\thanks{Yan-Ran Li is with the College of Computer Science and Software Engineering, Shenzhen University, China.}

}
\markboth{}
{Feng \MakeLowercase{\textit{et al.}}: Detect Faces Efficiently: A Survey and Evaluations}

\IEEEtitleabstractindextext{%
\begin{abstract}
Face detection is to search all the possible regions for faces in images and locate the faces if there are any. Many applications including face recognition, facial expression recognition, face tracking and head-pose estimation assume that both the location and the size of faces are known in the image. In recent decades, researchers have created many typical and efficient face detectors from the Viola-Jones face detector to current CNN-based ones. However, with the tremendous increase in images and videos with variations in face scale, appearance, expression, occlusion and pose, traditional face detectors are challenged to detect various "in the wild" faces. The emergence of deep learning techniques brought remarkable breakthroughs to face detection along with the price of a considerable increase in computation. This paper introduces representative deep learning-based methods and presents a deep and thorough analysis  in terms of accuracy and efficiency. We further compare and discuss the popular and challenging datasets and their evaluation metrics. A comprehensive comparison of several successful deep learning-based face detectors is conducted to uncover their efficiency using two metrics: FLOPs and latency. The paper can guide to choose appropriate face detectors for different applications and also to develop more efficient and accurate detectors.
\end{abstract}

}

\maketitle

\IEEEdisplaynontitleabstractindextext

%
\IEEEpeerreviewmaketitle


\IEEEraisesectionheading{\section{Introduction} \label{intro}}

\begin{figure}[htbp]
    \begin{center}
        \includegraphics[width=1.0\linewidth]{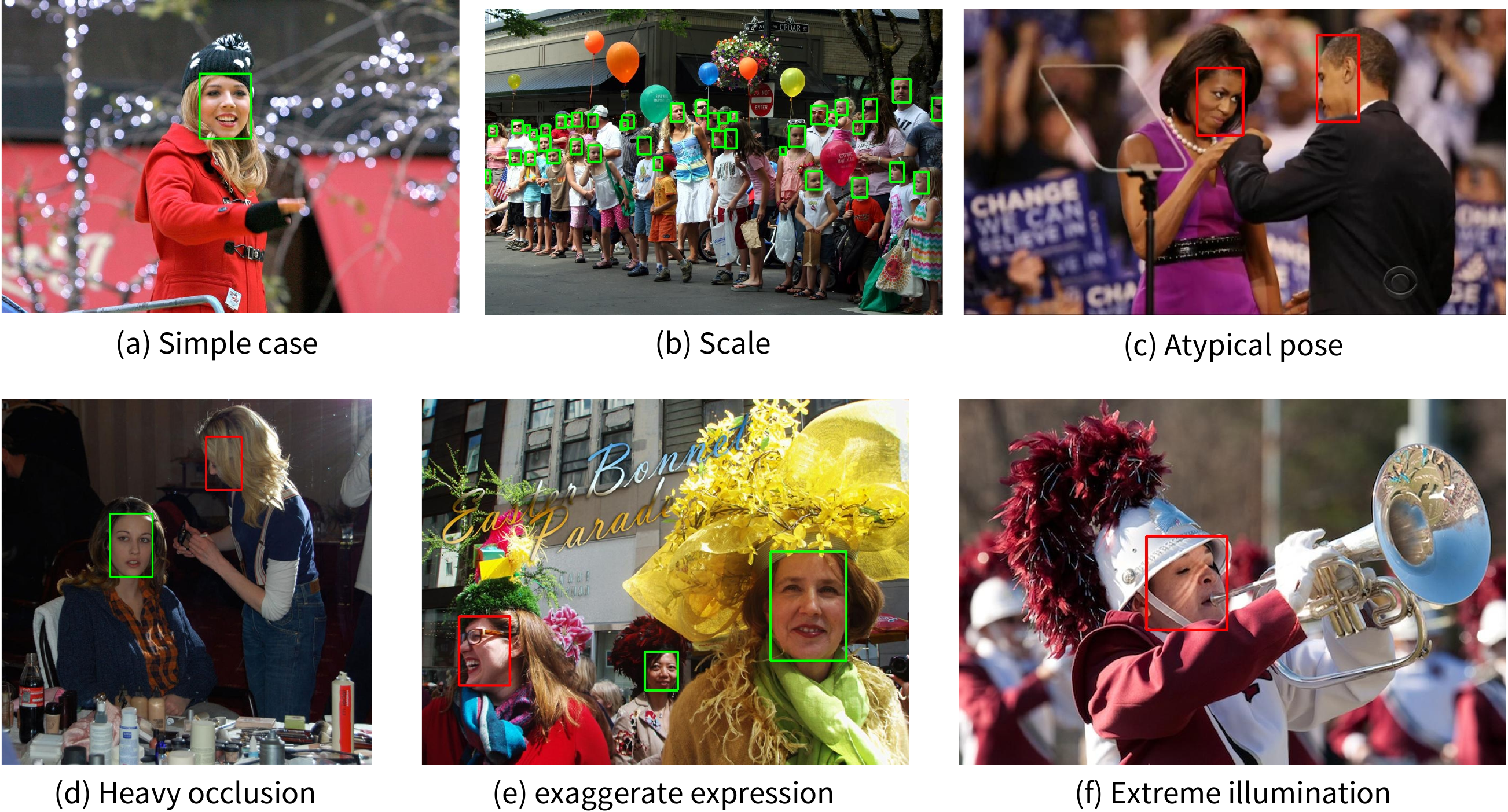}
    \end{center}
    \caption{Examples of face detection from WIDER Face~\cite{fd-widerface}. A simple case (a) where there is only one clear frontal face. Common variations are in scale (b), pose (c), occlusion (d), expression (e), illumination (f). Red boxes are faces in extreme conditions.}
    \label{fig:fd_intro}
\end{figure}

\IEEEPARstart{F}{ace} detection, one of the most popular, fundamental and practical tasks in computer vision, is to detect human faces from images and return the spatial locations of faces via bounding boxes~\cite{pascalvoc}, as shown in Fig.~\ref{fig:fd_intro}. Starting with the Viola-Jones (V-J) detector~\cite{Haar-like} in 2001, the solution to face detection has been significantly improved from handcrafting features such as Haar-like features~\cite{Haar-like}, to end-to-end convolutional neural networks (CNNs) for better feature extraction. Face detection is the first step for many face-related applications, such as face recognition, face tracking, facial expression recognition, facial landmarks detection and so on. Those technologies can achieve an overall better performance by faster and more accurate face detectors.

Before deep learning was employed for face detection, the cascaded AdaBoost classifier was the dominant method for face detection. Some algorithms were specifically designed for face detection by using some kinds of features, such as Haar-like features~\cite{Haar-like}, SURF~\cite{facesurf_li} and Multi-Block LBP~\cite{multi_block_LBP}. In recent years, deep learning has been proven to be more powerful for  feature extraction and helps to achieve very impressive accuracy on object detection. Numerous object detection deep models have been designed for generic object detection which is much more challenging than face detection. Therefore, many models from face detection are adopted from or inspired by models for generic object detection. We can train a deep face detector directly using Faster R-CNN~\cite{fasterrcnn}, YOLO~\cite{yolo} or SSD~\cite{ssd}, and much better detection results can be obtained than traditional cascaded classifiers. Some similar works can be found, such as Face R-CNN~\cite{fd-facercnn} and Face R-FCN~\cite{fd-facerfcn} which are modified and improved based on Faster R-CNN, R-FCN~\cite{rfcn} respectively. Additionally, some other detectors, such as  MTCNN~\cite{fd-mtcnn}, HR~\cite{fd-hr}, SSH~\cite{fd-ssh}, are originally designed for face detection. Some techniques in generic object detection have also been adapted into face detection, such as the multi-scale mechanism from SSD, the feature enhancement from FPN~\cite{fpn}, and the focal loss from RetineNet~\cite{focalloss} according to the special pattern of human faces for face detection. These techniques lead to the proposal of various outstanding face detectors such as S$^3$FD~\cite{fd-s3fd}, PyramidBox~\cite{fd-pyramidbox}, SRN~\cite{fd-srn}, DSFD~\cite{fd-dsfd}, and RetinaFace~\cite{retinaface}.

Face detection is sometimes considered as a solved problem because the average precision (AP) on many face detection datasets such as PASCAL Face~\cite{fd-pascalface}, AFW~\cite{fd-afw} and FDDB~\cite{fd-fddb}, has reached or exceeded 0.990 since 2017\footnote{State-of-the-art AP can be found in the official result pages of the datasets, and \url{https://paperswithcode.com/task/face-detection} which also collects results from published papers.}. On the most popular and challenging WIDER Face dataset~\cite{fd-widerface}, the AP has reached 0.921 even on the hard test set. 
\begin{figure}[htbp]
    \begin{center}
        \includegraphics[width=1.0\linewidth]{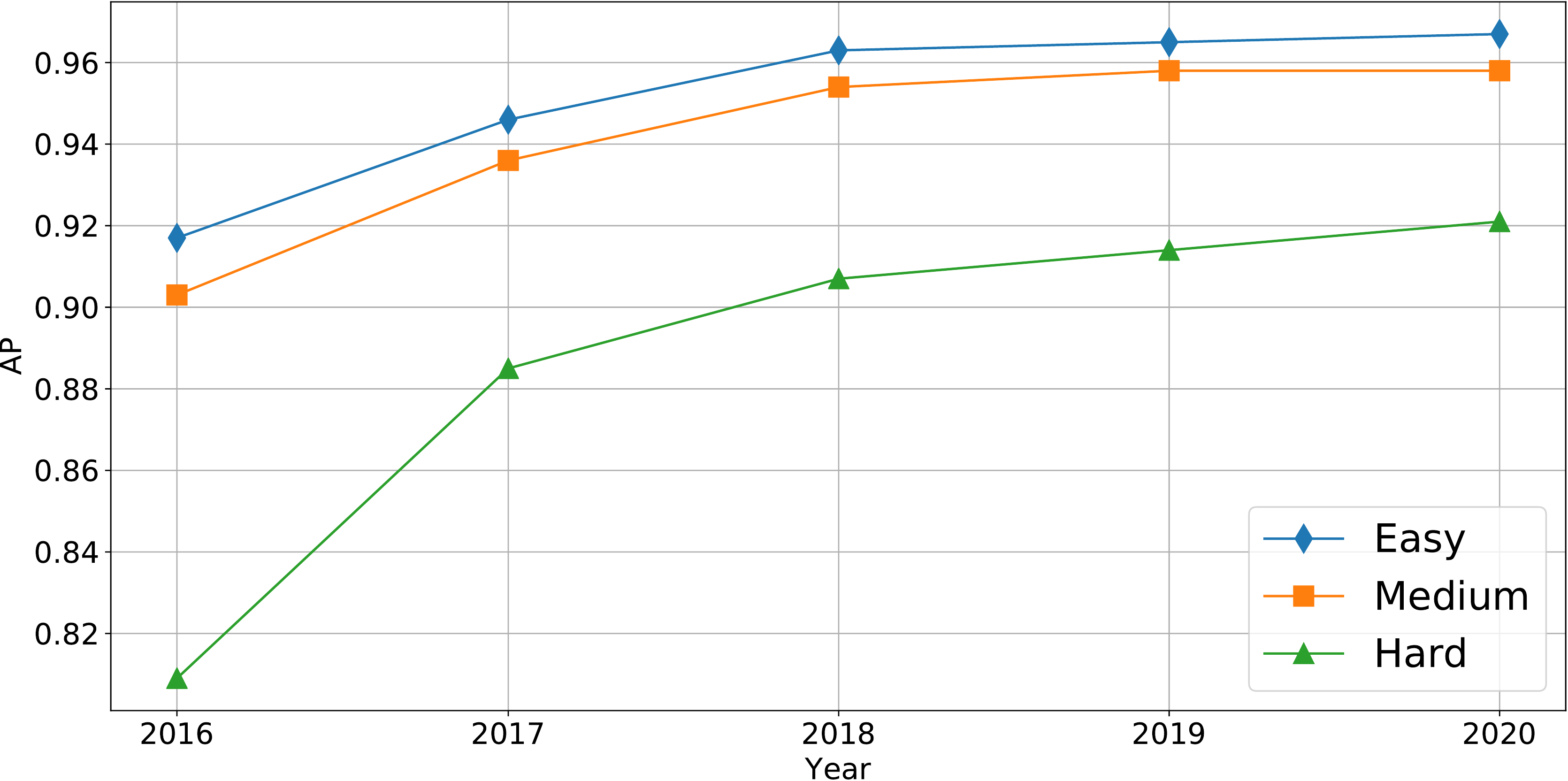}
    \end{center}
    \caption{
        The best AP on the easy, medium and hard subsets of WIDER Face~\cite{fd-widerface} test set in the recent years.
    }
    \label{fig:highest_ap_yearwise}
\end{figure}

But face detection is not a solved problem. If we observe the best results of each year in Fig.~\ref{fig:highest_ap_yearwise}, we can find the AP is still improving but slowly in recent 3 years. Therefore, with such near-to-saturated performance improvement, one question would be asked: If a tiny improvement is achieved by a much heavier deep model with great computational cost, will we consider the model is a good one? If we look slightly deeper into the implementation of some recent models, we can find that multiple scaling is heavily used in the evalutions on WIDERFACE benchmark. If we resize the input image with many different scales, such as 1/4, 1/2, 1, 3/2, 2, 4 and more, and feed all those resized images into a detector, the combined results will have a better AP, which in another word is achieved by the assembling and suppressing (NMS) the multi-scale outputs, and is independent to the backbone of the underlying face detector. We listed the scales used by some models in Table~\ref{tab:model_test_scales}. None of them tested an image using only one scale. It is a trend that more scales are used recently. There is a risk that multiple scales with a heavy computational cost are employed, and outstanding accuracy is claimed, which overshadows the performance gain the by the detector itself and the computational cost by such a multi-scale operation is not known. It is also worth noting that most benchmarks do not evaluate the computational cost. Most often, it is difficult for users to know by which the improvement is achieved, a better backbone technology or the follow-up computational-intensive multi-scale ensemble strategy?
\begin{table}[htbp]
    \centering
    \caption{
        Different models adopt different ranges and different presets of test scales. '0.25x' denotes shrinking the width and height by 0.25, and others follow. Specifically, 'Sx' and 'Ex' are shrinking and enlarging images accordingly, while 'Fx' is enlarging the image into a fixed size. Test image sizes stand for re-scaling the smaller side of the image to the given value, and the other side follows the same ratio.
    }
    \begin{tabular}{l|l}
    \hline
    Model      & Test image scales                                  \\ \hline
    HR,2017\cite{fd-hr}         & 0.25x, 0.5x, 1x, 2x                             \\
    S3FD,2017\cite{fd-s3fd}       & 0.5x, 1x, Sx, Ex                                \\
    SRN,2019\cite{fd-srn}        & 0.5x, 1x, 1.5x, 2.25x, Fx                       \\
    DSFD,2019\cite{fd-dsfd}       & 0.5x, 1x, 1.25x, 1.75x 2.25x, Sx, Ex            \\
    CSP,2019\cite{fd-csp}        & 0.25x, 0.5x, 0.75x, 1x, 1.25x, 1.5x, 1.75x, 2x  \\ 
    \hline
    Model & Test image sizes \\
    \hline
    SSH,2017\cite{fd-ssh}        & 500, 800, 1200, 1600                            \\
    SFA,2019\cite{SFA}        & 500, 600, 700, 800, 900, 1000, 1100, 1200, 1600 \\
    SHF,2020\cite{SHF}        & 100, 300, 600, 1000, 1400                       \\
    RetinaFace,~2020\cite{retinaface} & 500, 800, 1100, 1400, 1700                      \\ \hline
    \end{tabular}
    \label{tab:model_test_scales}
\end{table}

We do expect a perfect face detector which is robust and accurate even for some faces in extremely difficult conditions, while being extremely fast with low computational cost. However, we all know the \textit{no free lunch theorem}. Therefore, in this survey, we investigate the recent deep learning based face detection methods and evaluate them in terms of accuracy and computational cost. The main contributions are as follows.
\begin{enumerate}
    \item Different from previous face detection surveys~\cite{fd-survey2019, fd-survey2015, fd-survey2010, fd-survey2002, fd-survey2001} in which the content is mainly built on reviewing traditional methods, our survey focuses on deep learning-based face detectors. We have noted the existence of surveys~\cite{objectdetection-survey-zhao-2019, objectdetection-survey-zou-2019, objectdetection-survey-wu-2020} on deep learning; however, they focus on generic object detection, not specifically for face detection. In this paper, we provide a clear view of the path by which deep learning based face detection has evolved in
    recently years. 
    \item Accuracy and efficiency are both studied and analyzed in the paper. In addition to detailed introductions to deep learning based face detectors, some experiments are carried out to analyze different deep face detectors using different metrics. Some tricks to improve accuracy are also introduced. So the paper can help readers understand better how good accuracy and efficiency can be achieved. 
    \item With a focus on the efficiency of face detectors, comprehensive experiments are carried out to evaluate the accuracy and particularly efficiency of different face detectors. In addition to  latency, we also propose an accurate metric for the computational cost of a CNN model. It is \textbf{FL}oating point \textbf{OP}eration\textbf{s} (FLOPs) under certain rules. FLOPs is more neutral than latency which heavily depends on hardware and deep network structure. The code to compute the FLOPs has been released in \url{https://github.com/fengyuentau/PyTorch-FLOPs.git}.
\end{enumerate}

The rest of the paper is organized as follows. Some key challenges in face detection are summarized in Section~\ref{challenges}. In Section~\ref{review_arch}, we provide a roadmap to describe the development of deep learning-based face detection with detailed reviews. In Section~\ref{face_rep}, we review several fundamental subproblems including backbones, context modeling, the handling of face scale variations and proposal generation. Popular datasets for face detection and state-of-the-art performances are presented in Section~\ref{review_eval}. Section~\ref{comp} reveals the relation between computational cost and AP by conducting extensive experiments on several open-source one-stage face detectors. In addition, speed-focusing face detectors collected from Github are reviewed in Section~\ref{review_speed}. Finally, we conclude the paper with a discussion on future challenges in face detection in Section~\ref{conclusion}.

\section{Main Challenges} \label{challenges}

Most face-related applications need clear frontal faces. Detecting a clear frontal face is a relatively easy task. Some may argue that some faces are useless for the next step such as face recognition if the faces are tiny and with occlusion; but it is not. Effectively detecting any faces in extremely difficult conditions can greatly improve the perception capability of a computer but is still a challenging task. If a face is detected and evaluated as a bad quality sample, the subject can be suggested to be closer to the camera, or the camera can adjust automatically for a better image. Face detection is still a problem far from to be well solved. Many challenges do still exist.

\textbf{Accuracy-related challenges} are from face appearance and imaging conditions. In real-world scenes, there are many different kinds of face appearance, varying in different skin color, makeup, expression, wearing glasses or a mask and so on. In unconstrained environments, imaging a face can be impacted by various lighting, viewing angles and distances, backgrounds, and weather conditions. The face images will vary in illumination, pose, scale, occlusion, blur and distortion. The face samples in difficult conditions can be found in Fig.~\ref{fig:fd_intro}. There have been several datasets and competitions featuring face detection in unconstrained conditions, such as FDDB~\cite{fd-fddb}, WIDER Face~\cite{fd-widerface} and WIDER Face Challenge 2019~\footnote{\url{https://competitions.codalab.org/competitions/20146}}. More than 45\% of faces are smaller than $20\times 20$ pixels in WIDER. In most face-related applications, we seldom need small faces whose sizes are less than $20$. However, if we can detect small or even tiny faces, we can resize the original large images to smaller ones and send them to a face detector. Then, the computational cost can be greatly reduced since we only need to detect faces in smaller images. Therefore, a better accuracy sometimes also means a higher efficiency.

\textbf{Masked face detection} is becoming more important since people are wearing and will continuously wear masks to prevent COVID-19 in the next few years. Face-related applications did not consider this situation in the past. Wearing masks will reduce the detection accuracy obviously. Some masks are even printed with some logos or cartoon figures. All those can disrupt face detection. If a face has a mask and sunglasses at the same time, face detection will be even more difficult. Therefore, in the next few years, masked face detection should be explored and studied.

\textbf{Efficiency-related challenges} are brought by the great demands on edge devices. Since the increasing demands on edge devices, such as smartphones and intelligent CCTV cameras, massive amount of data is generated per day. We frequently take selfies, photos of others, long video meetings, etc. Modern CCTV cameras record 1080P videos constantly at 30 FPS. These result in a great demand for facial data analysis, and the data is considerable. In contrast, edge devices have limited computational capability, storage and battery life to run advanced deep learning-based algorithms. In this case, efficient face detection is essential for face applications on edge devices.

\begin{figure*}[htbp]
    \begin{center}
        \includegraphics[width=1.0\linewidth]{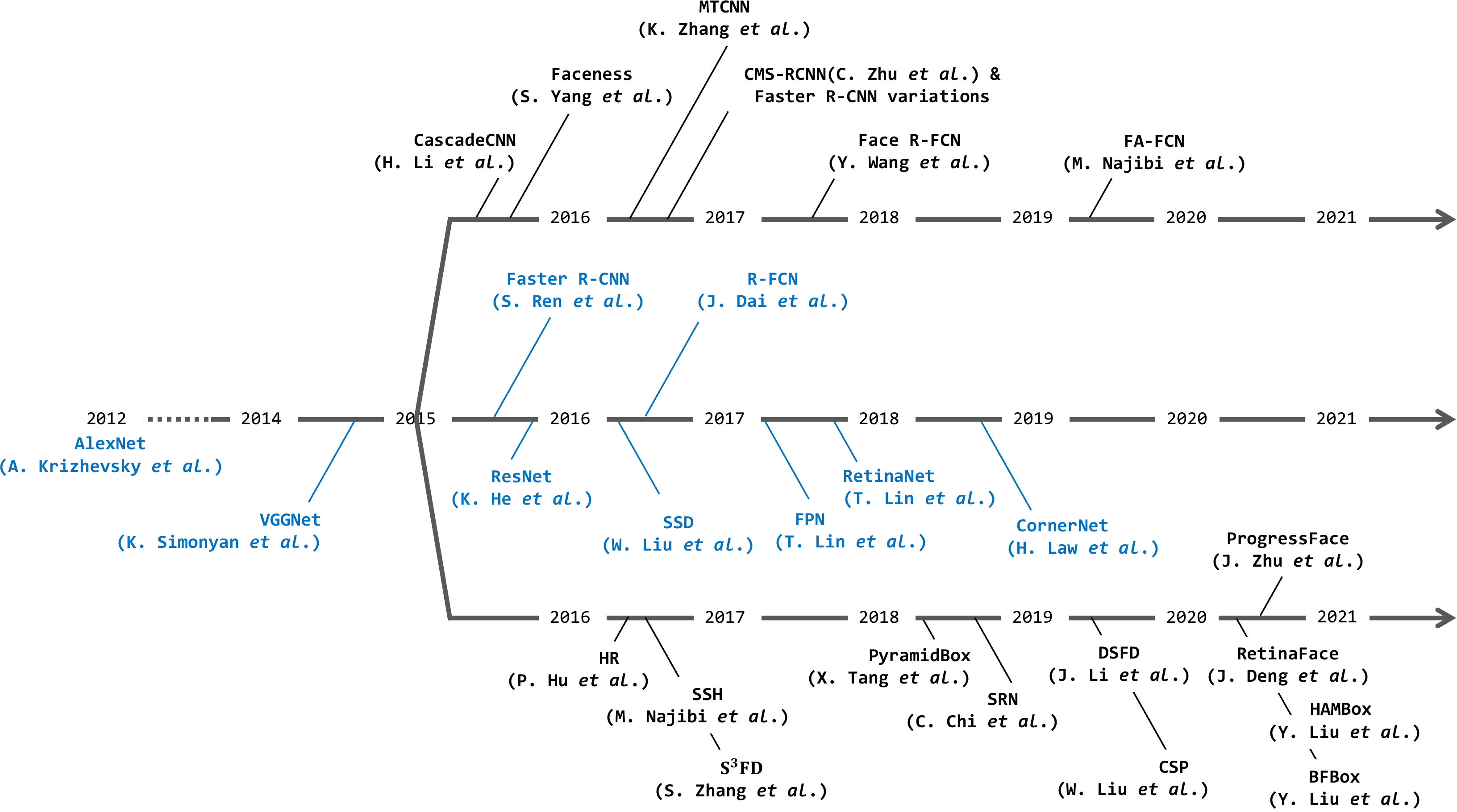}
    \end{center}
    \caption{
        Timeline of milestone face detectors~\cite{fd-cascadecnn, fd-faceness, fd-mtcnn, fd-cmsrcnn, fd-facerfcn, farpn, fd-hr, fd-ssh, fd-s3fd, fd-pyramidbox, fd-srn, fd-dsfd, fd-csp, retinaface, HAMBox, BFBox, ProgressFace}, and remarkable works from object recognition~\cite{bb-vgg, bb-resnet} and object detection~\cite{fasterrcnn, rfcn, ssd, fpn, focalloss, cornernet} (marked as blue, attached to the middle branch). Since the proposal of AlexNet~\cite{alexnet}, various face detection works inspired by deep learning techniques from object recognition and object detection were published in the 2012-post deep learning-based face detection era. The top branch is two/multi-stage face detectors, while the bottom branch is one-stage detectors, which has become the most popular network design adopted by researchers.
    }
    \label{fig:timeline}
\end{figure*}

\begin{figure*}[htbp]
    \begin{center}
        \includegraphics[width=1.0\linewidth]{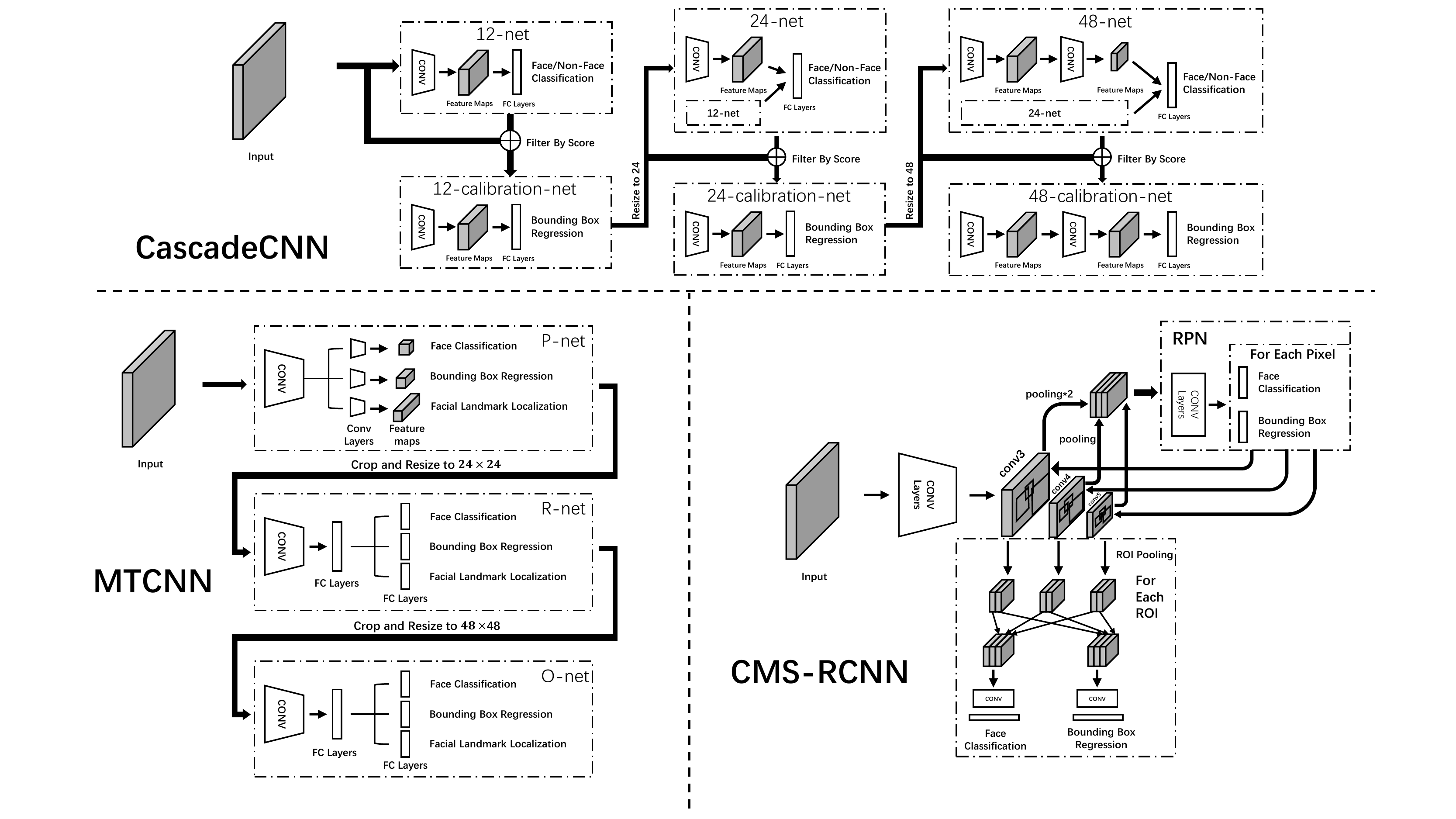}
    \end{center}
    \caption{
        Diagrams of milestone multi/two-stage face detectors~\cite{fd-cascadecnn, fd-mtcnn, fd-cmsrcnn, fd-cmsrcnn}. Others share similar architectures as the three.
    }
    \label{fig:multi-stage-archs}
\end{figure*}

\section{Face Detection Frameworks} \label{review_arch}

Before deep learning was used for face detection, cascaded AdaBoost-based classifiers were the most popular classifiers for face detection. The features used in AdaBoost were designed specifically for faces, not generic objects. For example, the Haar-like~\cite{Haar-like} feature can describe facial patterns of eyes, mouth and others. In recent years, facial features can be automatically learnt from data via deep learning techniques. Therefore, many deep learning-based face detectors are inspired by modern network architectures designed from object detection. Following the popular manner of organizing object detection frameworks, we organize deep learning-based face detectors into three main categories:
\begin{itemize}
    \item Multi-stage face detection frameworks. It is inspired by cascaded classifiers in face detection and is an early exploration of applying deep learning techniques to face detection.
    \item Two-stage face detection frameworks. The first stage generates some proposals, and the proposals are confirmed in the second stage. The efficiency should be better than multi-stage ones.
    \item One-stage face detection frameworks. Feature extraction and proposal generation are performed in a single unified network. These frameworks can be further categorized into anchor-based methods and anchor-free methods.
\end{itemize}

To show how the deep learning-based face detection evolves, milestone face detectors and some important object detectors are plotted in Fig.~\ref{fig:timeline}. The two-stage and multi-stage face detectors are on the top branch, and the single-stage ones are on the bottom branch. The generic object detectors are in the middle branch and in blue. A More detailed introduction of those detectors is provided in the following subsections.

\subsection{Multi-stage and Two-Stage Face Detectors}

In the early era when deep learning techniques entered face detection, face detectors were designed to have multiple stages, also known as the cascade structure which has been widely used in most early face detectors. With the remarkable breakthrough brought by Faster R-CNN~\cite{fasterrcnn}, some researchers turned to improve Faster R-CNN based on face data.

In the cascade structure, features are usually extracted and refined one or multiple times before being fed into classifiers and regressors, so as to reject most of the sliding windows to improve efficiency. As shown on the result page\footnote{{\url{http://vis-www.cs.umass.edu/fddb/results.html}}} of FDDB~\cite{fd-fddb}, Li et al. made an early attempt and proposed their CNN-based face detector, named \textbf{CascadedCNN}~\cite{fd-cascadecnn}. CascadeCNN consists of 3 stages of CNNs, as shown in Fig.~\ref{fig:multi-stage-archs}. Sliding windows are first resized to $12 \times 12$ pixels and fed into the shallow 12-net to reduce candidate windows by 90\%. The remaining windows are then processed by the 12-calibration-net to refine the size for face localization. Retained windows are then resized to $24 \times 24$ as the input for the combination of 24-net and 24-calibration-net, and so on for the next CNNs combination. CascadeCNN achieved state-of-the-art performance on AFW~\cite{fd-afw} and FDDB, while reaching a compelling speed of 14 FPS for for the typical $640 \times 480$ VGA images on a 2.0 GHz CPU. Another attempt at cascaded CNNs for face detection is the well-known \textbf{MTCNN}~\cite{fd-mtcnn} proposed by Zhang et al. MTCNN is composed of 3 subnetworks, which are P-Net for obtaining candidate facial windows, R-Net for rejecting false candidates and refining remaining candidates, O-Net for producing the final output with both face bounding boxes and landmarks in the multi-task manner. P-Net is a shallow fully convolutional network with 6 \texttt{CONV} layers, which can take images of any sizes as input. MTCNN was a great success with large and state-of-the-art advantages on WIDER Face~\cite{fd-widerface}, FDDB and AFW, while reaching 16 fps on a 2.6 GHz CPU.

In the object-detection-fashion two-stage network architectures, a region proposal network (RPN)~\cite{fasterrcnn} is required to generate object proposals. RPN can be considered as a straightforward classification CNN, which generates proposals based on the preset anchors on CNN features, filters out non objects and refines object proposals. However, as the CNNs shrink the image to extract features, the corresponding output features for tiny faces can be less than 1 pixel, making it insufficient to encode rich information. To address this problem, Zhu et al. proposed \textbf{CMS-RCNN}~\cite{fd-cmsrcnn}, which is equiped with a contextual multi-scale design for both RPN and final detection. As shown in Fig.~\ref{fig:multi-stage-archs}, multi-scale features from \textit{conv3}, \textit{conv4} and \textit{conv5} are concatenated by shrinking them into the same shape with \textit{conv5} as the input for RPN, so as to collect more information for tiny faces and also improve the localization capability from low-level layers. CMS-RCNN achieved an AP of 0.899, 0.874, 0.624 on the easy, medium and hard sets of the WIDER Face dataset respectively, outperforming MTCNN by 0.051(Easy), 0.049(Medium) and 0.016(Hard).

In addition to CMS-RCNN, there are others making improvements based on Faster R-CNN. \textbf{Bootstrapping Faster R-CNN}~\cite{Xiaomi} builds a training dataset by iteratively adding false positives from a model's output to optimize Faster R-CNN. \textbf{Face R-CNN}~\cite{fd-facercnn} adopts the same architecture as Faster R-CNN with center loss, online hard example mining and multi-scale training strategy. \textbf{FDNet}~\cite{fd-fdnet} exploits multi-scale training and testing and a vote-based NMS strategy on top of Faster R-CNN with a light-head design. Position-sensitive average pooling was proposed in \textbf{Face R-FCN}~\cite{fd-facerfcn} to assign different weights to different parts of the face based on R-FCN~\cite{rfcn}. With the improvements considering the special patterns of face data, these methods achieved better performance than their original version on the same WIDER Face dataset.

Whether it is the cascaded multi-stage or two-stage network design, its computation is heavily dependent on the number of faces in the image, the increase in which also increases proposals passed to the next stage in the interior of the network. Notably, the multi-scale test metric, which usually enlarges the images multiple times to make tiny faces detectable, can dramatically increase the computational cost on this basis. Considering that the number of faces in the image from the actual scene varies from one face in a selfie to many faces in a large group photo, we consider the robustness of cascade or two-stage networks in terms of runtime.


\subsection{One-Stage Face Detectors}


\begin{figure*}[t]
    \begin{center}
        \includegraphics[width=1.0\linewidth]{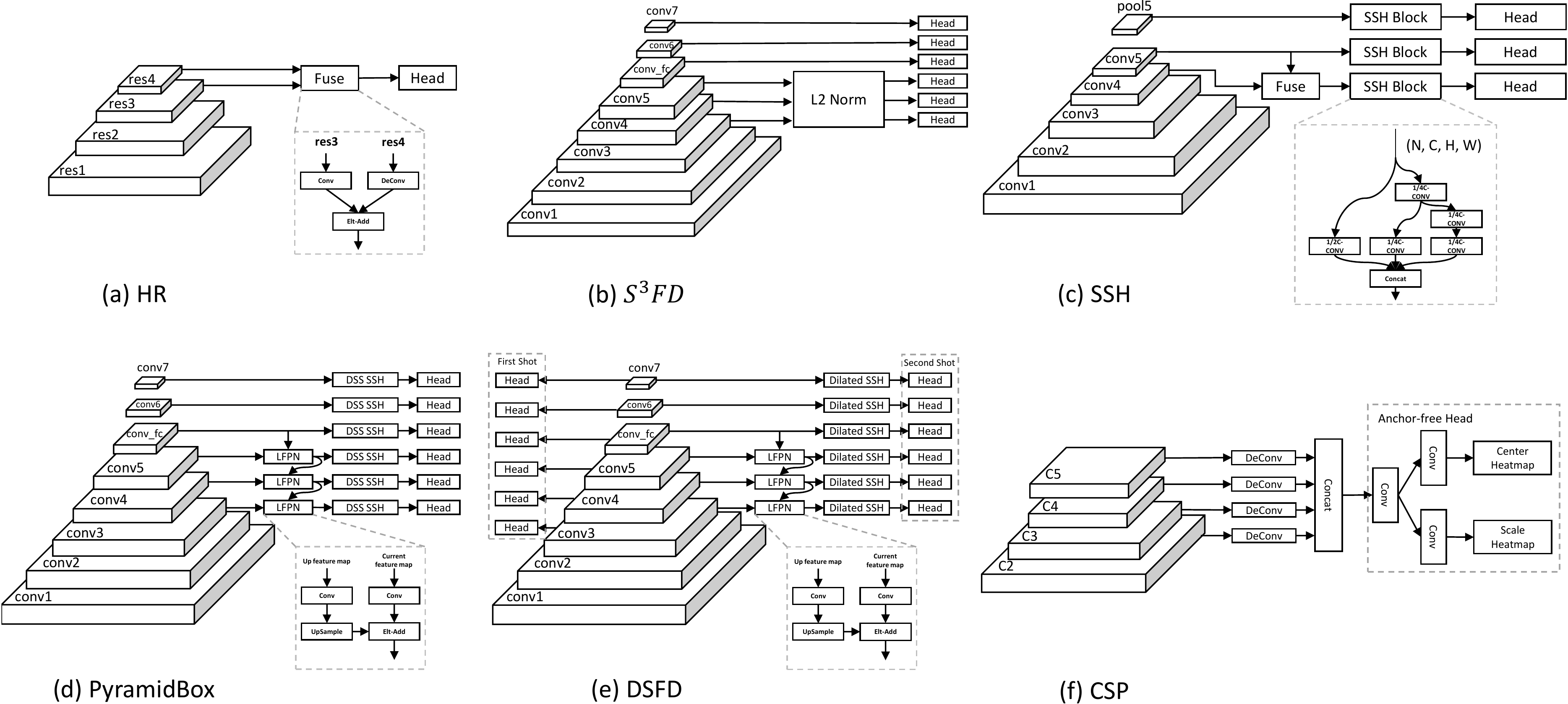}
    \end{center}
    \caption{
        Diagrams of milestone one-stage face detectors~\cite{fd-hr, fd-s3fd, fd-ssh, fd-pyramidbox, fd-dsfd, fd-csp}.
    }
    \label{fig:single-stage-archs}
\end{figure*}

In real-time face-related applications, face detection must be performed in real time. If the system is deployed on edge devices, the computing power is low. In those kinds of situations, one-stage face detectors are more suitable since their process time is stable regardless of how many faces there are in images. Different from the multi/two-stage detectors, the one-stage face detectors perform feature extraction, proposal generation and face detection in a single and unified convolutional neural network, whose runtime efficiency is independent of the number of faces. Dense anchors are designed to replace proposals in two-stage detectors~\cite{fd-ssh}. 
Starting from CornerNet~\cite{cornernet}, an increasing number of works use the anchor-free mechanism in their frameworks.


\textbf{HR}~\cite{fd-hr} proposed by Hu \textit{et al.} is one of the first to perform anchor-based face detection in a unified convolutional neural network. The backbone of HR is ResNet-101~\cite{bb-resnet} with layers truncated after \texttt{conv4\_5}. Early feature fusion on layers \texttt{conv3\_4} and \texttt{conv4\_5} is performed to encode context since high-resolution features are beneficial for small face detection. Through experiments on faces clustered into 25 scales, 25 anchors are defined for 2X, 1X and .5X inputs, to achieve the best performance of three input scales. HR outperformed CMS-RCNN~\cite{fd-cmsrcnn} by 0.199 on the WIDER Face validation hard set, and more importantly, the run-time of HR is independent of the number of faces in the image, while CMS-RCNN's linearly scale up with the number of faces.

Different from HR, \textbf{SSH}~\cite{fd-ssh} attempts to detect faces at different scales on different levels of features, as shown in Fig.~\ref{fig:single-stage-archs}. Taking VGG-16~\cite{bb-vgg} as the backbone, SSH detects faces on the enhanced features from \texttt{conv4\_3}, \texttt{conv5\_3} and \texttt{pool5} for small, medium and large faces respectively. SSH introduces a module (SSH module) that greatly enriches receptive fields to better model the context of faces. The SSH module is widely adopted by later works~\cite{fd-pyramidbox, fd-dsfd, retinaface, HAMBox}, which turns out to be efficient for performance boosting.

Since \textbf{S$^3$FD}~\cite{fd-s3fd}, many one-stage face detectors~\cite{fd-pyramidbox, fd-srn, fd-dsfd, fd-csp, retinaface, HAMBox, BFBox, ProgressFace} fully utilize multi-scale features attempting to achieve scale-invariant face detection. S$^3$FD extends the headless VGG-16~\cite{bb-vgg} with more convolutional layers, whose stride gradually doubles from 4 to 128 pixels, so as to cover a larger range of face scales. \textbf{PyramidBox}~\cite{fd-pyramidbox} adopts the same backbone as S$^3$FD, integrates FPN~\cite{fpn} to fuse adjacent-level features for semantic enhancement, and improves the SSH module with wider and deeper convolutional layers inspired by Inception-ResNet~\cite{Inception-v4} and DSSD~\cite{dssd}. \textbf{DSFD}~\cite{fd-dsfd} also inherits the backbone from S$^3$FD, but enhances the multi-scale features by the Feature Enhance Module (FEM), so that detection can be made on two shots - one from non-enhanced multi-scale features, and the other from the enhanced features. The same scale features from the second shot have larger RFs than those from the first shot, but also have smaller RFs than the next-level features from the first shot, indicating that the face scales are split more refined across these multi-scale detection layers. Similarly, \textbf{SRN}~\cite{fd-srn} has a dual-shot networks but is trained differently on multi-scale features: low-level features need two-step classification to refine, since they have higher resolution and contribute the vast majority of anchors and also negative samples; additionally, high-level features have lower resolution which is worth two-step regression using the Cascade R-CNN~\cite{cascadercnn} to have more accurate bounding boxes.

There are also some significant anchor-based methods using the FPN~\cite{fpn} as the backbone. \textbf{RetinaFace} adds one more pyramid layer on top of the FPN and replaces CONV layers with the deformable convolution network (DCN)~\cite{DCN, DCNv2} within FPN's lateral connections and context module. RetinaFace models a face in three ways: a 3D mesh (1k points), a 5-landmark mask (5 points), and a bounding box (2 points). Cascade regression~\cite{cascadercnn} is employed with multi-task loss in RetinaFace to achieve better localization. Instead of using the handcrafting structures, Liu \textit{et al} proposed \textbf{BFBox}, which explores face-appropriate FPN architectures using the successful Neural Architecture Search (NAS). Liu decouples FPN as the backbone and FPN connections, the former of which can be replaced by VGG~\cite{bb-vgg}, ResNet~\cite{bb-resnet} or the backbone from NAS, and the latter of which can be top-down, bottom-up or cross-level fusion from NAS.


Since the proposal of CornerNet~\cite{cornernet} back in 2018, which directly predicts the top-left and bottom-right points of bounding boxes instead of relying on prior anchors, many explorations~\cite{fcos, zhoucenternet, extremenet, reppoints} have been made to remodel object detection more semantically using the anchor-free design. \textbf{CSP} models a face bounding box as a center point and the scale of the box as shown in Fig.~\ref{fig:single-stage-archs}. CSP takes multi-scale features from the modified ResNet-50~\cite{bb-resnet}, and concatenates them to take the advantage of rich global and local information for detection heads using transpose convolution layers. In particular, the anchor-free detection head can also be an enhancement module for anchor-based heads. \textbf{ProgressFace}~\cite{ProgressFace} appends an anchor-free module to provide more positive anchors for the highest resolution feature maps in FPN, so as to reduce the imbalance of positive and negative samples for small faces.

One-stage frameworks are popular on face detection in recent years for the following three reasons. (a) The runtime of one-stage face detectors is independent of the number of faces in an image by design. Therefore, it enhances the robustness of runtime efficiency. (b) It is computationally efficient and straightforward for one-stage detectors to reach near scale invariance by contextual modeling and multi-scale feature sampling. (c) Face detection is a relatively less complex task than general object detection. This means that innovations and advanced network designs in object detection can be quickly adjusted to face detection by considering the special pattern of faces.


\section{Face Representation} \label{face_rep}

The key idea of face detection has never changed whether it is in the traditional era or deep learning era. It finds the common patterns of all faces in the dataset. 
In the traditional era, many of handcrafted features, such as SIFT~\cite{SIFT}, Haar~\cite{Haar-like} and HOG~\cite{HOG}, are employed to extract local features from the image, which are aggregated by approaches such as AdaBoost for the higher-level representation of faces. 

Different from traditional methods, which require rich prior knowledge to design handcrafted features, deep convolutional neural networks can directly learn even more powerful features from face images. A deep learning-based face detection model can be considered as two parts: a CNN backbone and several detection branches. Starting from some popular CNN backbones, the feature extraction methods that can handle face scale invariance are introduced as well as several strategies to generate proposals for face detection.

\subsection{Popular CNN Backbones} \label{backbones}

In most deep face detectors there is a CNN backbone for feature extraction. Some popular backbone networks are listed in Table~\ref{tab:cnn-backbones}. They are VGG-16 from the VGGNet~\cite{bb-vgg} series, ResNet-50/101/152 from the ResNet~\cite{bb-resnet} series, and MobileNet~\cite{MobileNet-v1}. The models are powerful and can achieve good accuracy on face detection, but they are a little heavy.
\begin{table}[htbp]
    \centering
    \caption{
        CNN backbones commonly used by modern deep learning-based face detectors. FC layers of these CNNs are ignored when calculating '\#CONV Layers', '\#Params' and 'FLOPs'. The input size for calculating 'FLOPs' is $224 \times 224$. The calculation of FLOPs is discussed in Section~\ref{comp}. 'Top-1 Error' refers to the performance on the ImangeNet~\cite{imagenet} validation set. Note that 9 of the 20 \texttt{CONV} layers in MobileNet~\cite{MobileNet-v1} are depth-wise.
    }
    \begin{tabular}{|c|c|c|c|c|}
    \hline
    \begin{tabular}[c]{@{}c@{}}CNN\\ Backbones\end{tabular} & \begin{tabular}[c]{@{}c@{}}\#CONV\\ Layers\end{tabular} & \begin{tabular}[c]{@{}c@{}}\#Params\\ ($\times 10^6$)\end{tabular} & \begin{tabular}[c]{@{}c@{}}FLOPs\\ ($\times 10^9$)\end{tabular} & \begin{tabular}[c]{@{}c@{}}Top-1\\ Error\end{tabular} \\ \hline
    VGG-16           & 13   & 14.36  & 30.72  & 28.07\%  \\ \hline
    ResNet-50    & 52   & 23.45  & 8.25   & 22.85\%  \\ \hline
    ResNet-101    & 136  & 42.39  & 15.72  & 21.75\%  \\ \hline
    ResNet-152    & 188  & 56.87  & 23.19  & 21.43\%  \\ \hline
    MobileNet  & 20   & 3.22   & 1.28   & 29.40\%  \\ \hline
    \end{tabular}
    \label{tab:cnn-backbones}
\end{table}

\begin{figure}[htbp]
    \begin{center}
        \includegraphics[width=1.0\linewidth]{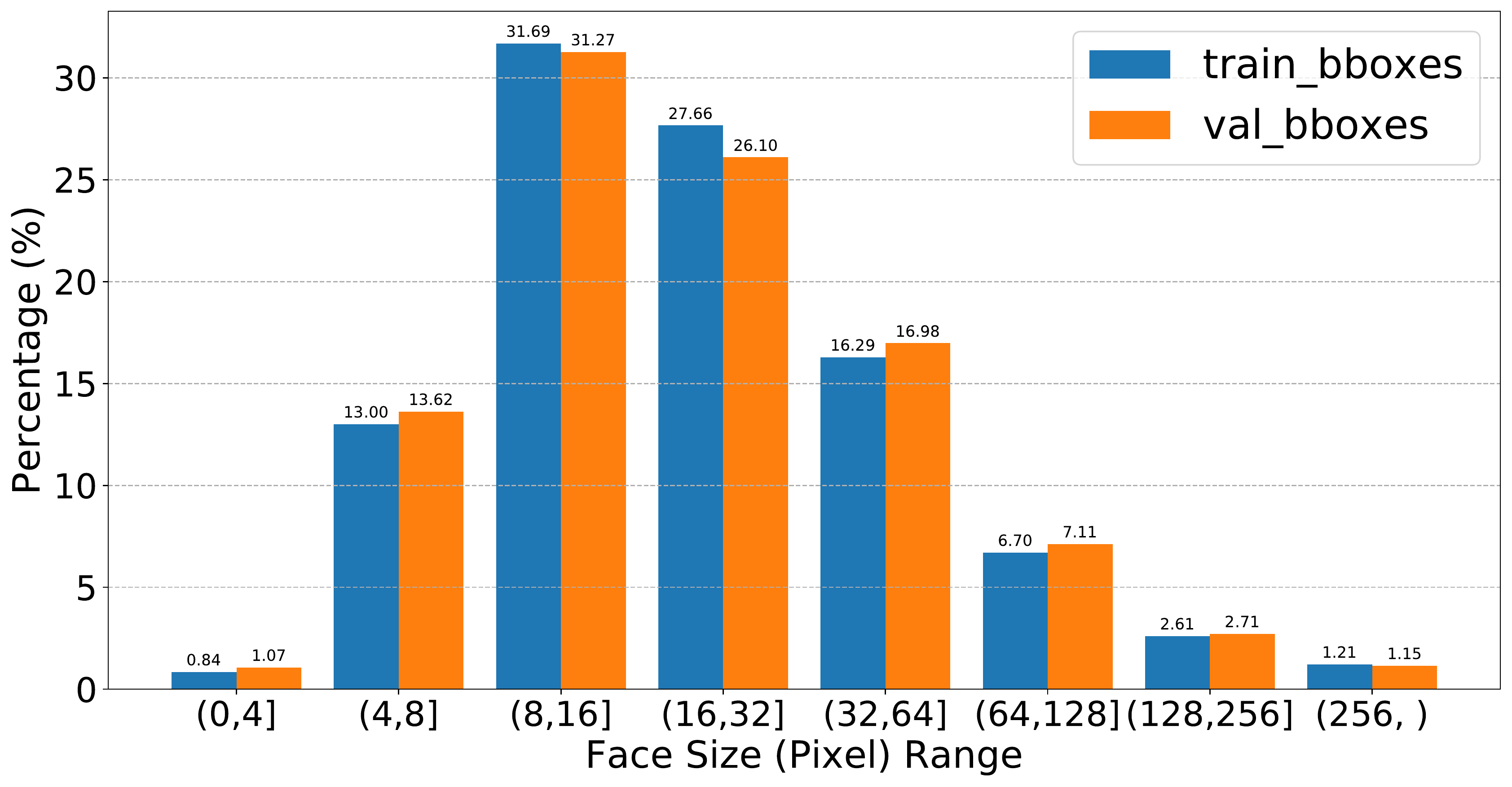}
    \end{center}
    \caption{
        The distribution of face scales on WIDER Face~\cite{fd-widerface} dataset. 
    }
    \label{fig:Scale_dis}
\end{figure}

Early attempts on deep learning-based face detection were cascaded structures that did not take the above CNN architectures. Even some simple structured CNN is much more computational heavy than AdaBoost, cascaded CNN is computational heavy also. With breakthroughs in object detection, some of the techniques have been borrowed and applied on face detection. VGG-16~\cite{bb-vgg} has 13 \texttt{CONV} layers, which is the first choice for the baseline backbones for many face detectors, such as SSH~\cite{fd-ssh}, S$^3$FD~\cite{fd-s3fd} and PyramidBox~\cite{fd-pyramidbox}. Performance improvements can easily be obtained by simply swapping the backbone from VGG-16 to ResNet-50/101/152~\cite{bb-resnet}, as shown in ~\cite{fd-dsfd}. Since state of the arts have achieved AP \textgreater 0.900 even on WIDER Face hard sets, it is common for recent face detectors~\cite{fd-dsfd, ProgressFace, tinaface} to equipe with a deeper and wider backbone for higher AP, such as the ResNet-152 and ResNets with FPN~\cite{fpn} connections. Liu \textit{et al.} employs Neural Architecture Search (NAS) to search face-appropriate backbones and FPN connections.

One of the most inexpensive choices is ResNet-50 which is listed in Table~\ref{tab:cnn-backbones}, which has less parameters and less FLOPs, while achieving very similar performance compared to deeper nets. Another choice for state-of-the-art face detectors to reach real-time speed is to change the backbone to MobileNet~\cite{MobileNet-v1}, which has similar performance to VGG-16 but one order of magnitude less in '\#Params' and FLOPs.

\begin{figure}[htbp]
    \begin{center}
        \includegraphics[width=1.0\linewidth]{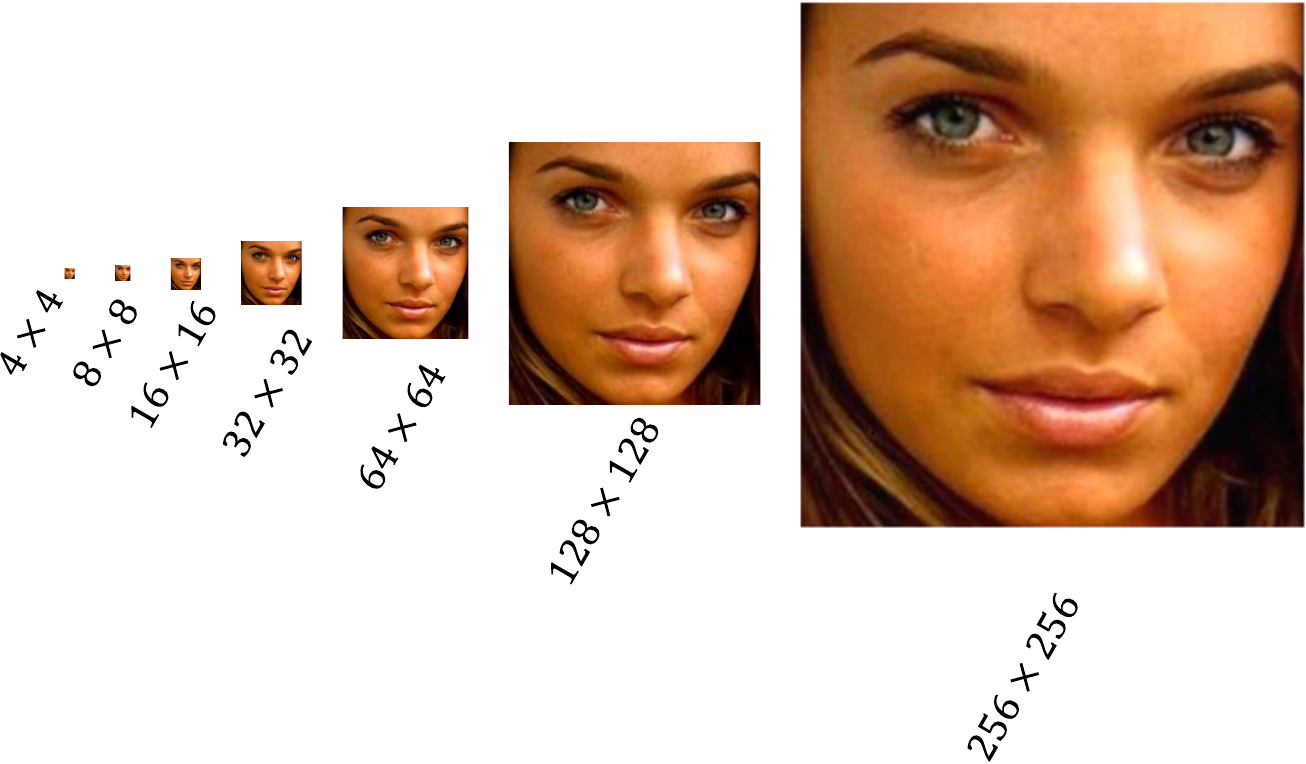}
    \end{center}
    \caption{
        A face in different scales. Could you tell the images of sizes $4 \times 4$, $8 \times 8$ contain a face?
    }
    \label{fig:face_in_scales}
\end{figure}

\subsection{Towards Face Scale Invariance}

One of the major challenges for face detection is the large span at face scales. As statistics shown in Fig.~\ref{fig:Scale_dis}, there are 157,025 and 39,123 face bounding boxes in the train and validation set respectively, both of which have more than 45\% of face bounding boxes are $16 \times 16$ and smaller, and a non-negligible ~1\% are $256 \times 256$ and larger. We choose these scales to perform clustering to match the strides of feature maps selected for detection; for example there is only 1 pixel in the feature maps of stride 4 for encoding a face of size equal to or less than $4 \times 4$. We also present the visual differences among scales in Fig.~\ref{fig:face_in_scales}. It is challenging even for humans to tell whether the image of size $16 \times 16$ contains a face. In the following, we describe the mechanism of face detectors towards face scale invariance even with tiny faces.

Most of the modern face detectors are anchor-based. Anchors are predefined boxes of different scales and aspect ratios attached to each pixel in the feature maps, which serve as the proposal to match with the ground truth faces. More details about anchors are provided in Section~\ref{proposal_generation}. As \cite{fd-s3fd} noted, since the predefined anchor scales are discrete while the face scales in the wild change continuously, outer faces whose scales are distributed away from anchor scales cannot match enough anchors. It will result in a low recall rate. A simple solution for a trained face detector is to perform multi-scale test on an image pyramid, which is built by progressively resizing the original image. It is equal to re-scale faces and hopefully brings outer faces back into the detectable range of scales. This solution does not require retraining the detector, but it may come with a sharp increase in redundant computation, since there is no certain answer to how deep the pyramid we should build to match with the certain extent of scale invariance of a trained CNN.

Another better solution to face scale invariance is to make full use of the feature maps produced in CNNs. One can easily observe that the layers of standard CNN backbones gradually decrease in size. The subsampling of these layers naturally builds up a pyramid with different strides and receptive fields (RFs). It produces multi-scale feature maps. In general, high-level feature maps produced by later layers with large RFs are encoded with strong semantic information, and lead to its robustness to variations such as illumination, rotation and occlusion. Low-level feature maps produced by early layers with small RFs are less sensitive to semantics, but have high resolution and rich details, which are beneficial for localization. To take both the advantages, a number of methods are proposed, which can be categorized into \textbf{modeling context}, \textbf{detecting on a feature pyramid}, and \textbf{predicting face scales}.

\textbf{Modeling context:} Additional context is essential for detecting faces, especially for detecting small ones. HR\cite{fd-hr} shows that context modeling by fusing feature maps of different scales can dramatically improve the accuracy of detecting small faces. Following a similar fusion strategy as HR, \cite{SHF} detects on three different dilated \texttt{CONV} branches, aiming to enlarge RF without too much increase in computation. \cite{fd-cmsrcnn} downsamples feature maps of strides 4 and 8 to concatenate with those of stride 16, so as to improve the capability of the RPN to produce proposals for faces at different scales. SSH~\cite{fd-ssh} exploits an approach similar to Inception~\cite{Inception-V1}, which concatenates the output from three \texttt{CONV} branches that have $3 \times 3$, $5 \times 5$ and $7 \times 7$ filters respectively. PyramidBox~\cite{fd-pyramidbox} first adopts an FPN~\cite{fpn} module to build up context and is further enhanced by deeper and wider SSH modules. \cite{fd-dsfd} improves the SSH module by replacing \texttt{CONV} layers with dilated \texttt{CONV} layers. \cite{fd-csp} upsamples feature maps of strides 8, 16 to concatenate with those of stride 4, which is fed to an FCN to produce center, scale and offset heatmaps. The fusion of feature maps encodes rich semantics from high-level feature maps with rich geometric information from low-level feature maps, based on which the detectors can improve their capability of localization and classification towards face scale invariance. Meanwhile, the fusion of feature maps also introduces more layers, such as \texttt{CONV} and \texttt{POOL} to adjust scales and channels, which creates additional computational overhead.

\textbf{Detecting on a feature pyramid}: Inspired by SSD~\cite{ssd}, a majority of recent approaches, such as \cite{fd-ssh,fd-s3fd,fd-pyramidbox,fd-srn,fd-dsfd,retinaface}, detect at multiple feature maps of different scales respectively, and combine detection results. It is considered to be an effective method for weighing between speed and accuracy. SSD~\cite{ssd} puts default boxes on each pixel of the feature maps from 6 detection layers that have strides of 8, 16, 32, 64 and 128. Sharing a similar CNN backbone with SSD, \cite{fd-s3fd,fd-pyramidbox} detect on a wider range of layers, which have strides gradually doubling from 4 to 128 pixels. SRN~\cite{fd-srn} and DSFD~\cite{fd-dsfd} introduce the two-stream mechanism, which detects on both the detection layers from the backbone and extra layers applied on the detection layers for feature enhancement. Different from subsampling on more layers, \cite{fd-ssh,SFA,retinaface} detects only at the last three level feature maps, which are enhanced by their context modeling methods. By detecting on a feature pyramid, detection layers are implicitly trained to be sensitive to different scales, while it also leads to an increase in model size and redundant computation, since the dense sampling may cause some duplicate results from adjacent-level layers.

\textbf{Predicting face scales}: To eliminate the redundancy from pyramids, several approaches~\cite{SAFD,RSA,S2AP} predict the face scales before making a detection. \cite{SAFD} first generates a global face scale histogram from the input image by the Scale Proposal Network (SPN), which is trained with image-level ground truth histogram vectors and without face location information. A sparse image pyramid is built according to the output histogram, so as to have faces rescaled to the detectable range of the later single-scale RPN. Similarly, \cite{RSA} detects on a feature pyramid without unnecessary scales, which is built by using the scale histogram to a sequential ResNet~\cite{bb-resnet} blocks that can downsample feature maps recursively. \cite{S2AP} predicts not only face scales but also face locations by a shallow ResNet18~\cite{bb-resnet} with scale attention and spatial attention attached, named S$^2$AP. S$^2$AP generates a 60-channel feature map, meaning face scales are mapped to 60 bins, each of which is a spatial heatmap that has high response to its responsible face scale. With the 60-channel feature maps, it is possible to decrease the unnecessary computation with the low-response channels and the low-response spatial areas by a masked convolution.

\subsection{Proposal Generation} \label{proposal_generation}


Faces in the wild can be of any possible locations and scales in the image. The general pipeline for most of the early successful face detectors, is to first generate proposals in the sliding-window manner, extract features from the windows using handcrafted descriptors~\cite{Haar-like,fd-afw,6619289,1410446} or CNNs~\cite{fd-cascadecnn,fd-mtcnn}, and finally apply face classifiers. However, inspired by RPN~\cite{fasterrcnn} and SSD~\cite{ssd}, modern anchor-based face detectors generate proposals by applying $k$ anchor boxes on each pixel of the extracted CNN features. Specifically, 3 scales and 3 aspect ratios are used in Faster R-CNN~\cite{fasterrcnn}, yielding $k=9$ anchors on each pixel of the feature maps. Moreover, the detection layer takes the same feature maps as input, yielding $4k$ outputs encoding the coordinates for $k$ anchor boxes from the regressor and $2k$ outputs for face scores from the classifier.

Considering that most of the face boxes are near square, modern face detectors tend to set the aspect ratio of anchors to 1, while the scales depends. HR~\cite{fd-hr} defines 25 scales so as to match the cluster results on the WIDER Face~\cite{fd-widerface} training set. S$^3$FD assigns the anchor scale of 4 times the stride of the current layer to keep anchor sizes smaller than effective receptive fields~\cite{ERF} and ensure the same density of different scale anchors on the image. PyramidBox~\cite{fd-pyramidbox} introduces PyramidAnchors, which generates a group of anchors with larger regions corresponding to a face, such as head and body boxes, to have more context to help detect faces. In ~\cite{zcc}, extra shifted anchors are added to increase the anchor sample density, and significantly increased the average IoU between anchors and small faces. GroupSampling~\cite{GroupSampling} assigns anchors of different scales only on the bottom pyramid layer of FPN~\cite{fpn}, but it groups all training samples according to the anchor scales, and randomly samples from groups to ensure the positive and negative sample ratios between groups are the same.



\section{Datasets and Evaluation} \label{review_eval}

\begin{table*}[htbp]
    \centering
    \caption{Comparison of currently accessible face detection datasets, listed in the order of publication or started year. Note that UCCS~\cite{uccs} and WILDEST Face~\cite{wildestface} are not included because their data is not currently available. 'Blur', 'App.', 'Ill.', 'Occ.', 'Pose' in the 'Variations' columns denote blur, appearance, illumination, occlusion and pose respectively.}
    \label{tab:dataset-stats}
    \begin{tabular}{|c|c|c|c|c|c|c|c|ccccc|}
    \hline
    \multirow{2}{*}{Dataset} & \multirow{2}{*}{\#Images} & \multirow{2}{*}{\#Faces} & \multirow{2}{*}{\begin{tabular}[c]{@{}c@{}}\#Faces\\ Per Image\end{tabular}} & \multirow{2}{*}{\begin{tabular}[c]{@{}c@{}} AVG Resolution\\  ($W \times H$)\end{tabular}} & \multicolumn{3}{c|}{Split} & \multicolumn{5}{c|}{ Variations} \\ \cline{6-13} & &                          & & & Train   & Val    & Test    & Blur & App. & Ill. & Occ. & Pose \\ \hline
    FDDB~\cite{fd-fddb}  & 2,845  & 5,171  & 1.8  & $377 \times 399$ & - & - & 100\% & \checkmark & & &  \checkmark &  \checkmark \\ \hline
    AFW~\cite{fd-afw}    & 205    & 468    & 2.3  &  $1491 \times 1235$ & - & - & 100\% & &  \checkmark & & &  \checkmark \\ \hline
    PASCAL Face~\cite{fd-pascalface} & 851 & 1,335 & 1.5 & - & - & - & 100\% & & & & & \\ \hline
    MALF~\cite{fd-malf} & 5,250 & 11,931 & 2.2 & - & - & - & 100\%   & &  \checkmark & &  \checkmark &  \checkmark \\ \hline
    WIDER Face~\cite{fd-widerface} & 32,203 & 393,703 & 12.2 &  $1024 \times 888$ & 40\%    & 10\%   & 50\%    &  \checkmark &  \checkmark &  \checkmark &  \checkmark &  \checkmark \\ \hline
    MAFA~\cite{mafa} & 30,811 & 39,485 & 1.2 &  $516 \times 512$ & 85\% & - & 15\% & &  \checkmark & &  \checkmark & \\ \hline
    IJB-A~\cite{ijba} & 48,378 & 497,819 & 10.2 &  $1796 \times 1474$ & 50\% & - & 50\% & &  \checkmark &  \checkmark &  \checkmark &  \checkmark \\ \hline
    IJB-B~\cite{ijbb} & 76,824 & 135,518 & 1.7 &  $894 \times 599$ & - & - & 100\% & &  \checkmark &  \checkmark &  \checkmark &  \checkmark \\ \hline
    IJB-C~\cite{ijbc} & 138,836 & 272,335 & 1.9 &  $1010 \times 671$ & - & - & 100\% & &  \checkmark &  \checkmark &  \checkmark &  \checkmark \\ \hline
    4K-Face~\cite{4kface} & 5,102 & 35,217 & 6.9 &  $3840 \times 2160$ & - & - & 100\% & & & & & \\ \hline
    UFDD~\cite{ufdd} & 6,425 & 10,897 & 1.6 &  $1024 \times 774$ & - & - & 100\% &  \checkmark & &  \checkmark & & \\ \hline
    DARK Face~\cite{darkface} & 6,000 & 43,849 & 7.3 &  $1080 \times 720$ & 100\% & - & - & & &  \checkmark & & \\ \hline
    \end{tabular}
\end{table*}

To evaluate different face detection algorithms, datasets are needed. There have been several public datasets, which are FDDB~\cite{fd-fddb}, AFW~\cite{fd-afw}, PASCAL Face~\cite{fd-pascalface}, MALF~\cite{fd-malf}, WIDER Face~\cite{fd-widerface}, MAFA~\cite{mafa}, 4K-Face~\cite{4kface}, UFDD~\cite{ufdd} and DARK Face~\cite{darkface}. These datasets all consist of colored images from real-life scenes. Different datasets may utilize different evaluation criterion. In Section~\ref{datasets}, we present overviews of different datasets covering some statistics such as the number of images and faces, the source of images, the rules of labeling and challenges brought by the dataset. A detailed analysis of the face detection evaluation criterion is also included in Section~\ref{eval-criterion}. Detection results on the datasets are provided and analyzed in Section~\ref{datasets-results}.

\subsection{Datasets} \label{datasets}

Some essential statistics of currently accessible datasets are summarized in Table~\ref{tab:dataset-stats} including the total number of images and faces, faces per image, how the data was splitted different sets, etc. More details are introduced in the following part.

\textbf{FDDB}\footnote{\url{http://vis-www.cs.umass.edu/fddb/}}~\cite{fd-fddb} is short for \textbf{F}ace \textbf{D}etection \textbf{D}ataset and \textbf{B}enchmark, which has been one of the most popular datasets for 
face detector evaluation since its publication in 2010. The images of FDDB were collected from Yahoo! News, 2,845 of which were selected after filtering out duplicate data. Faces were excluded with these factors, (a) height or width less than 20 pixels, (b) the two eyes being non-visible, (c) the angle between the nose and the ray from the camera to the head being less than 90 degrees, (d) failure estimation on position, size or orientation of faces by a human. This led to 5,171 faces left, which were annotated by drawing elliptical face regions covering from the forehead to the chin vertically, and the left cheek to the right cheek horizontally. FDDB helped advance uncontrained face detection in terms of the robustness of expression, pose, scale and occlusion. However, its images can be heavily biased toward celebrity faces since they were collected from the news. It is also worth noting that although the elliptical style of the face label adopted by FDDB is closer to human cognition, it is not adopted by later datasets and deep learning-based face detectors, which favor the bounding box style with a relatively easier method for defining positive/negative samples by calculating the Intersection over Union (IoU).

Zhu et al. built an annotated faces in-the-wild (\textbf{AFW}\footnote{\url{http://www.cs.cmu.edu/~deva/papers/face/index.html}}) dataset~\cite{fd-afw} by randomly sampling images with at least one large face from Flickr. 468 faces were annotated from 205 images, each of which is labeled with a bounding box and 6 landmarks. \textbf{PASCAL Face}\footnote{\url{http://host.robots.ox.ac.uk/pascal/VOC/}}~\cite{fd-pascalface} was contructed by selecting 851 images from the PASCAL VOC~\cite{pascalvoc} test set with 1,335 faces annotated. Since the two datasets were built to help evaluate the face detectors proposed by ~\cite{fd-afw} and ~\cite{pascalvoc}, they only contain a few hundred images, resulting in limited variations in face appearance and background.

Yang et al. created the \textbf{M}ulti-\textbf{A}ttribute \textbf{L}abelled \textbf{F}aces~\cite{fd-malf} (\textbf{MALF}\footnote{\url{http://www.cbsr.ia.ac.cn/faceevaluation/}}) dataset for fine-grained evaluation on face detection in the wild. The MALF dataset contains 5,250 images from Flickr and Baidu Search with 11,931 faces labeled, which is an evidently larger dataset than FDDB, AFW and PASCAL Face. The faces in MALF were annotated by drawing axis-aligned square bounding boxes, attempting to contain a complete face with the nose in the center of the bounding box. This may introduce noise for training face detectors since a square bounding box containing a 90-degree side faces can have over half of its content being cluttered background. In addition to labeling faces, some attributes were also annotated, such as gender, pose and occlusion.

In 2016, \textbf{WIDER Face}\footnote{\url{http://shuoyang1213.me/WIDERFACE/}}~\cite{fd-widerface} was released, which has been the most popular and widely used face detection benchmark. The images in WIDER Face were collected from popular search engines for predefined event categories following LSCOM~\cite{lscom} and examined manually to filter out similar images and images without faces, resulting in 32,203 images in total for 61 event categories, which were split into 3 subsets for training, validation testing set. To keep large variations in scale, occlusion and pose, the annotation was performed following two main policies: (a) a bounding box should tightly contain the forehead, chin and cheek and is drew for each recognizable face and (b) an estimated bounding box should be drawn for an occluded face, producing 393,703 annotated faces in total. The number of faces per image reaches 12.2 and 50\% of the faces are of height between 10-50 pixels. WIDER Face outnumbers other datasets in Table~\ref{tab:dataset-stats} by a large margin. It means WIDER Face pays never-seen-before attention to small faces detection by providing a large number of images with the densest small faces for training, validation and testing. Furthermore, the authors of WIDER Face defined 'easy', 'medium' and 'hard' levels for the validation and test sets based on the detection rate of EdgeBox~\cite{edgebox}. It offers a much more detailed and fine-grained evaluation for face detectors. Hence, the WIDER Face dataset greatly advances the researches of CNN based face detectors, especially the multi-scale CNN designs and utilization of context.

The last four datasets listed in Table~\ref{tab:dataset-stats} are less generic than those reviewed above, and focus on face detection in specified and different aspects. The \textbf{MAFA}\footnote{\url{http://www.escience.cn/people/geshiming/mafa.html}}~\cite{mafa} dataset focuses on masked face detection, containing 30,811 images with 39,485 masked faces labeled. In addition to the location of eyes and masks, the orientation of the face, the occlusion degree and the mask type were also annotated for each face. The IJB series\footnote{\url{https://www.nist.gov/programs-projects/face-challenges}}~\cite{ijba, ijbb, ijbc} were collected for multiple tasks, including face detection, verification, identification, and identity clustering. The IJB-C is the combination of IJB-A and IJB-B with some new face data. \textbf{4K-Face}\footnote{\url{https://github.com/Megvii-BaseDetection/4K-Face}}~\cite{4kface} was built for the evaluation of large face detection, and contains 5,102 4K-resolution images with 35,217 large faces (\textgreater512 pixels). \textbf{UFDD}\footnote{\url{https://ufdd.info}}~\cite{ufdd} provides a test set with 6,425 images and 10,897 faces in the variation of different weather conditions and degradtion such as lens impediments. \textbf{DARK Face}\footnote{\url{https://flyywh.github.io/CVPRW2019LowLight/}}~\cite{darkface} concentrates on face detection in low light conditions, and provides 6,000 low-light images for training dark face detector. Since the images are captured in real-world nighttime scenes such as streets, each image in DARK Face contains 7.3 faces on average which is 
relatively dense.

\begin{figure*}[htbp]
    \centering
    \subfloat[Discontinuous ROC curves]{
        \includegraphics[width=0.45\linewidth]{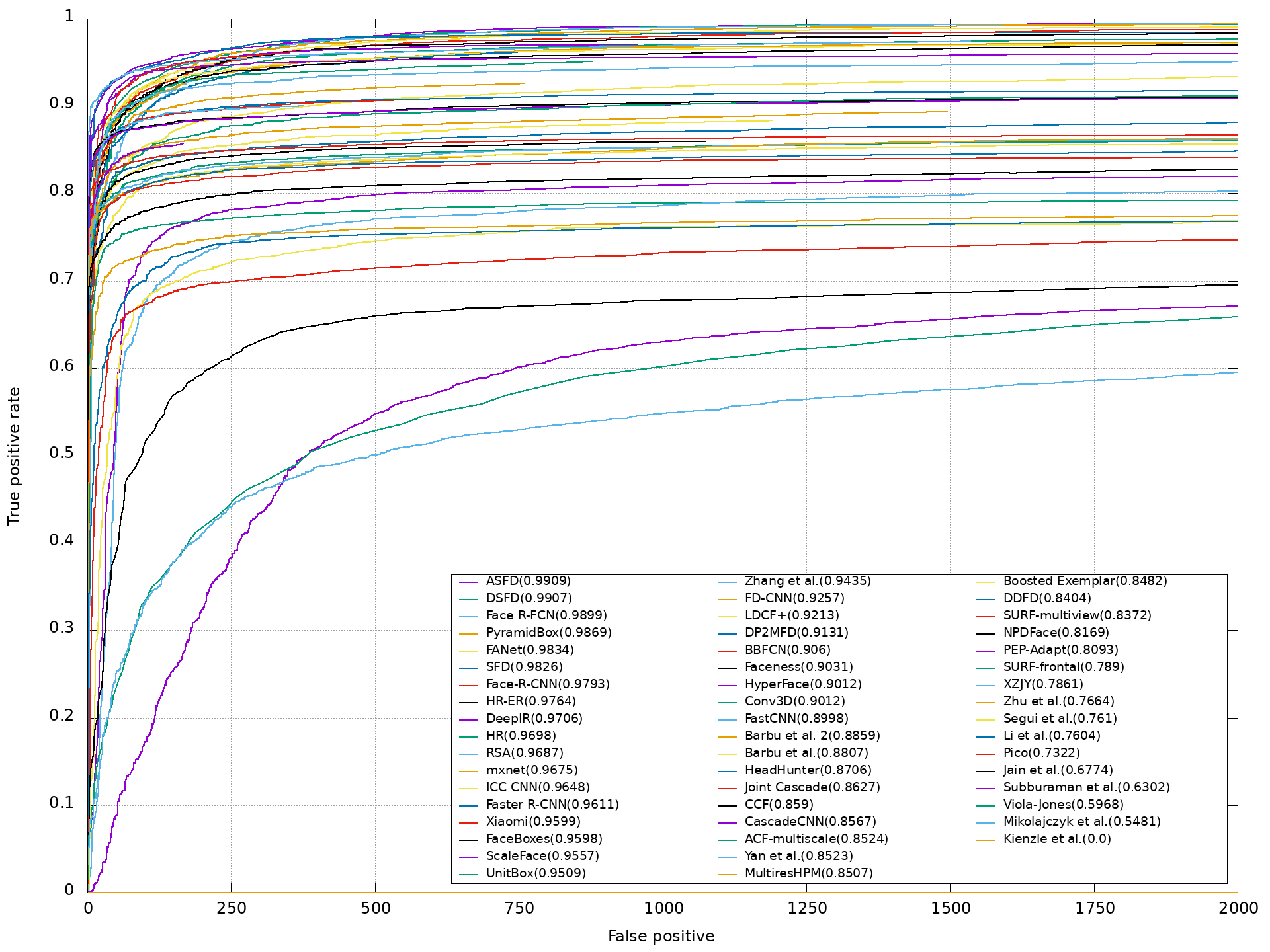}
    }
    \qquad
    \subfloat[Continuous ROC curves]{
        \includegraphics[width=0.45\linewidth]{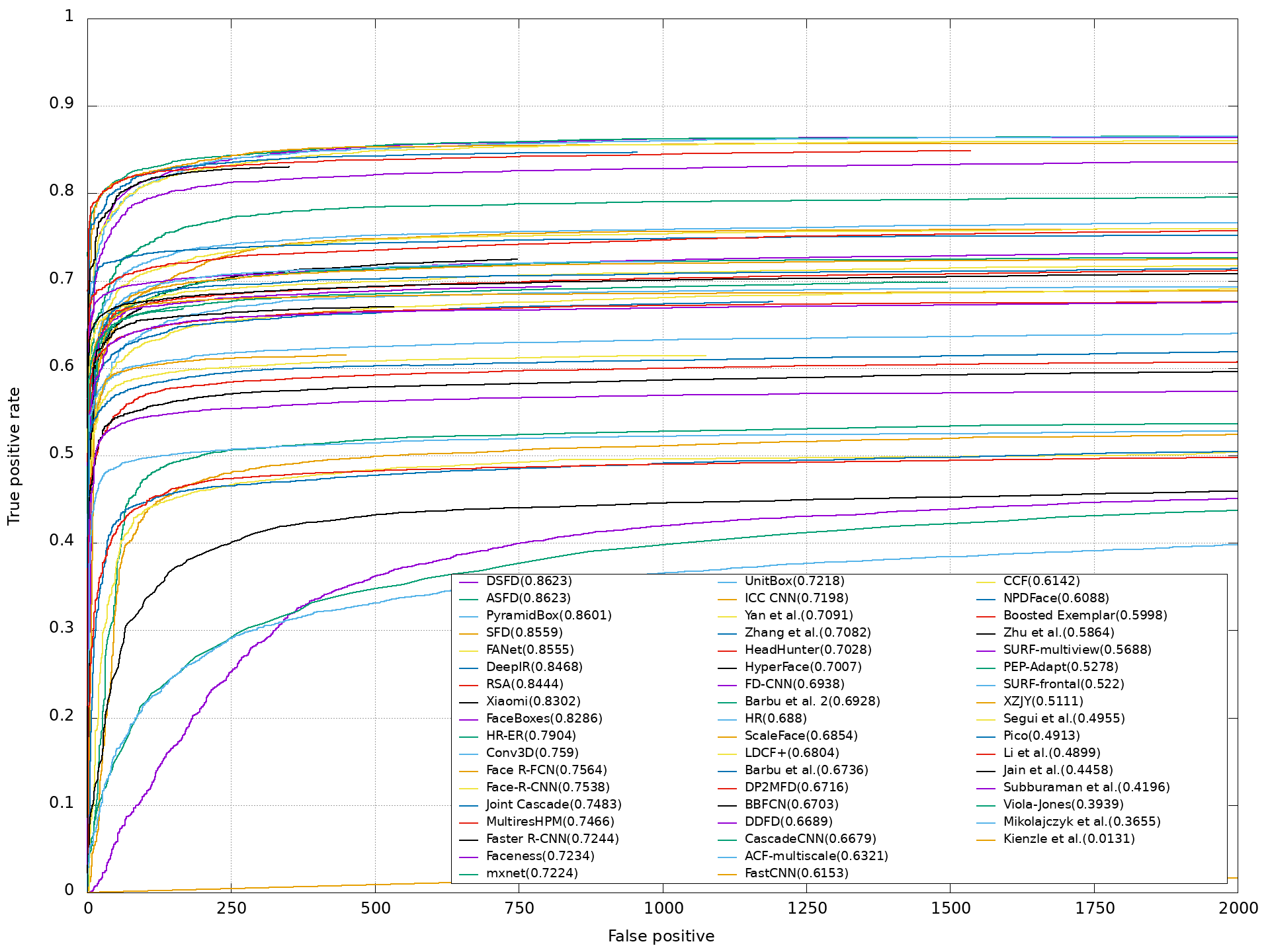}
    }
    \caption{
        The results on the FDDB dataset, which are from the result page of FDDB {\url{http://vis-www.cs.umass.edu/fddb/results.html}}.
    }
    \label{fig:fddb_sota}
\end{figure*}

\begin{figure*}[htbp]
    \subfloat[WIDER Face Validation Set]{
        \includegraphics[width=1.0\linewidth]{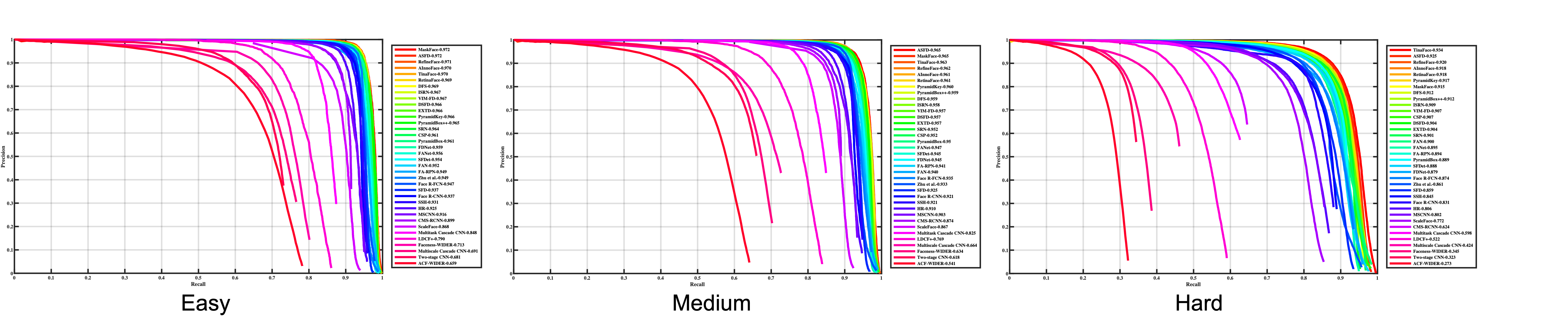}
    }
    \\
    \subfloat[WIDER Face Test Set]{
        \includegraphics[width=1.0\linewidth]{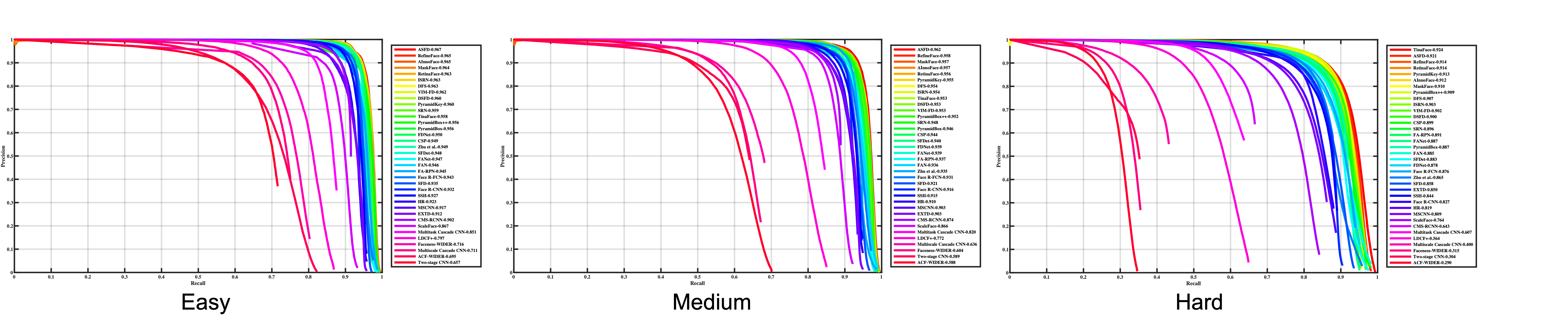}
    }
    \caption{
        The results on the WIDER Face validation and test sets. The figures are from WIDER face homepage \url{http://shuoyang1213.me/WIDERFACE/}.
    }
    \label{fig:wider_sota}
\end{figure*}

\subsection{Accuracy Evaluation Criterion} \label{eval-criterion}

There are mainly two accuracy evaluation criteria adopted by the datasets reviewed above, one of which is the receiver operating characteristic (ROC) curve obtained by plotting the true positive rate (TPR) against false positives such as those adopted by FDDB~\cite{fd-fddb}, MALF~\cite{fd-malf}, UCCS~\cite{uccs} and IJB~\cite{ijbc}, the other of which is the most popular evaluation criterion from PASCAL VOC~\cite{pascalvoc} by plotting the precision against recall while calculating average precision (AP), such as those adopted by AFW~\cite{fd-afw}, PASCAL Face~\cite{fd-pascalface}, WIDER Face~\cite{fd-widerface}, MAFA~\cite{mafa}, 4K-Face~\cite{4kface}, UFDD~\cite{ufdd}, DARK Face~\cite{darkface} and Wildest Face~\cite{wildestface}. Since these two kinds of evaluation criterion are two different methods for revealing the performance of detectors under the same calculation of the confusion matrix~\footnote{\url{https://en.wikipedia.org/wiki/Confusion_matrix}}, we choose the most popular evaluation criteria AP calculated from the precision-again-recall curve in the paper.

To get a precision-again-recall curve, the confusion matrix, which is to define the true positives (TP), false positives (FP), false negatives (FN) and true negatives (TN) from the detection and ground truths, should be firstly calculated. A true positive is a detection result matched with a ground truth; otherwise, it is a false positive. The unmatched ground truths are defined as the false negatives. True negatives are not applied here since the background can be a large part of the image. To define whether two regions are matched or not, the commonly used intersection over 
union (IoU), also known as the Jaccard overlap, is applied:
\begin{equation} \label{eq:ioudef}
    IoU = \frac{area(P) \cap area(GT)}{area(P) \cup area(GT)}
\end{equation}
where $P$ is the predicted region, and $GT$ is the ground truth region. In a widely used setting, the IoU threshold is set to 0.5, meaning if the IoU of a predicted region and a ground truth region is greater than or equal to 0.5, the predicted region is marked as matched and thus a true positive, otherwise it is a false positive.

After determining true or false positives for each detection, the next step is to calculate the precision and recall from the detection result list sorted by score in descending order to plot the precision-against-recall curve. A granular confidence gap can be defined to sample more precision and recall, but for a simple explanation, we define the gap as a detection result. In $n$th sampling, we calculate the precision and recall from the top-$n$ detection results:
\begin{align} \label{eq:prec-recall-def}
    Precision_{n} = \frac{TP_{n}}{TP_{n} + FP_{n}}
    \\
    Recall_{n} = \frac{TP_{n}}{TP_{n} + FN_{n}}
\end{align}
where $TP_{n}$, $FP_{n}$ and $FN_{n}$ are true positives, false positives and false negatives from the top-$n$ results respectively. Let us say we have 1,000 detection results; then, we have 1,000 pairs of $(recall_i, precision_i)$ which are enough for plotting the curve.

We can compute the area under the precision-against-recall curve, which is AP, to represent the overall performance of a face detector. Under the single IoU threshold setting of 0.5 in WIDER Face evaluation, the top AP for the hard test subset of WIDER reached 0.924. In the WIDER Face Challenge 2019 which uses the same data as the WIDER Face dataset but evaluates face detectors in 10 IoU thresholds of 0.50:0.05:0.95, the top average AP reaches 0.5756.

\subsection{Results on Accuracy} \label{datasets-results}

To understand the progress in recent years on face detection, the results of different datasets are collected from their official homepages. Because of space limitations, only the results from the two most popular datasets are listed. They are Fig.~\ref{fig:fddb_sota} for FDDB~\cite{fd-fddb} and Fig.~\ref{fig:wider_sota} for WIDER Face~\cite{fd-widerface}.
The FDDB results since 2004 are listed. The current ROC curves are much better than those in the past. This means that the detection accuracy is much higher than in the past. The true positive rate is reaching 1.0. If you look into the samples in FDDB, you can find there are some tiny and blur faces in the ground truth data. Sometimes it is hard to decide whether they should be faces, even by humans. Therefore, we can say that the current detectors achieve perfect accuracy on FDDB, and almost all faces can be detected.

The WIDER face is newer, larger and more challenging than FDDB. Most recent face detectors have been tested with it. From Fig.~\ref{fig:wider_sota}, it can be found that the accuracy is also very high even on the hard set. The improvement on mAP is not so obvious now. The mAP is almost saturated similar to FDDB.

We must note that the current benchmarks, regardless of FDDB, WIDER or others, only evaluate the accuracy of detection and do not evaluate efficiency. If two detectors achieve similar mAP, but the computational cost of one is just half of another, surely we will think the detector with half computational cost is better than another. Since the accuracy metric is almost saturated, it is time to include efficiency in the evaluation.

\section{Evaluation of Computational Cost} \label{comp}

\begin{table*}[htbp]
    \caption{Equations of FLOPs calculation of different layers.}
    \label{tab:flops-calc}
    \centering
    \begin{tabularx}{\textwidth}{|c|c|X|}
    \hline
    NN layers & FLOPs & Explanation \\ \hline
    Conv &
    $C_{out} H_{out} W_{out} (2 C_{in} K^2 - 1)$ &
    \begin{tabular}{l}
        For each element in the output tensor, there are $C_inK^2$ multiplications between the \\
        kernels and sliding windows, and $C_inK^2-1$ additions to sum up. If bias is used, 1 \\
        FLOPs should be added to the FLOPs calculation of each element.~\cite{conv-eq}
    \end{tabular} \\ \hline
    Max Pool &
    $K^2 C_{out} H_{out} W_{out}$ &
    \begin{tabular}{l}
        For each element in the output tensor, we consider the worst situation where every \\
        element in the kernel requires a comparison with each other.
    \end{tabular}\\ \hline
    ReLU &
    $2 C_{out} H_{out} W_{out}$ &
    \begin{tabular}{l}
        ReLU is usually implemented as $x*(x>0)$, which is much faster than directly \\
        comparing $x$ with $0$. We consider a comparison as 1 FLOPs for simplicity.
    \end{tabular}\\ \hline
    Batch Norm &
    $6 C_{out} H_{out} W_{out}$ &
    \begin{tabular}{l}
        As~\cite{batchnorm} stated, the variances and means are fixed during inference. Therefore, 6 FLOPs\\
        is accounted for applying the linear transform to each element.
    \end{tabular}\\ \hline
    L2-Norm &
    $3 C_{out} H_{out} W_{out}$ &
    \begin{tabular}{l}
        The L2-norm layer was proposed by~\cite{l2norm} to help features of late fusion work well, \\ 
        which is defined as $L_2-norm(x)=\frac{x}{||x||_2}=\frac{x}{\sqrt{\sum^d_{i=1}|x_i|^2}}$, where $d$ usually stands for \\
        channels. It takes approximately $2CHW$ FLOPs to calculate the $L_2$ norm \\ 
        channel-wisely and $CHW$ FLOPs to perform $L_2$ norm element-wisely.
    \end{tabular} \\ \hline
    Bilinear Upsample &
    $19 C_{out} H_{out} W_{out}$ &
    \begin{tabular}{l}
        The definition of bilinear upsampling~\footnote{\url{https://en.wikipedia.org/wiki/Bilinear_interpolation}} contains 9 non-duplicate additions \\
        and subtractions and 10 multiplications/divisions for calculating one element in the \\
        output.
    \end{tabular} \\ \hline
    Sigmoid &
    $3 C_{out} H_{out} W_{out}$ &
    \begin{tabular}{l}
        The definition of sigmoid~\footnote{\url{https://en.wikipedia.org/wiki/Sigmoid_function}} contains 1 exponentiation, 1 addition and 1 division to \\
        calculate one element in the output.
    \end{tabular} \\ \hline
    Softmax &
    $3E$ &
    \begin{tabular}{l}
        $E$ denotes the total number of elements in the output tensor. It takes approximately \\
        $2E$ FLOPs to calculate the sum of the exponentiation of each element in different \\
        channels, and $E$ FLOPs to calculate the final result.
    \end{tabular} \\ \hline
    \end{tabularx}
\end{table*}

Deep learning techniques have brought momentous improvement to face detection, and can detect faces more robustly in unconstrained environments. Most of the recent works train and test their models on WIDER Face~\cite{fd-widerface}. As shown in Fig.~\ref{fig:highest_ap_yearwise}, we can find a large AP leap from 2016 to 2017. However, the line has been flat since 2017. If we look deep into the official releasing code of recent works, it can be easily found that newer models tend to use larger scales and a wider range of scales as shown in Table~\ref{tab:fd-test-scales}. These test scales are usually not mentioned in the papers, but can lead to a non-negligibly great increase in computational cost just for slightly boosting the AP. We may even question: Is the AP improved by a better algorithm or the usage of a wider range of test scales?

\subsection{Rules of FLOPs Calculation} \label{sec:rule-flops}

\textbf{What kind of models are we going to re-evaluate?} First, the models must be open-source at least with the release of its test code and a trained model. We do not re-implement the methods since we want to ensure that the accuracy should be 100\% the same as the original authors claimed. Additionally, it is essential for us to choose one-stage models, as their FLOPs are independent of the number of faces in the images, and they have been the most studied frameworks in recent years. Third, we mainly choose the models from the WIDER Face result page for fair comparisons.

\begin{table*}[htbp]
    \centering
    \caption{Test scales used by open-source one-stage face detectors~\cite{fd-hr, fd-ssh, fd-s3fd, fd-pyramidbox, fd-srn, fd-dsfd, fd-csp}. Note that the double check marks denote image flipping vertically in addition to the image at the current scale. SSH shrinks and enlarges images to several preset fixed sizes. Since S$^3$FD, two adaptive test scales are used to save GPU memory, one of which is "S" for adaptive shrinking, the other of which is "E" for recursively adaptive enlarging. Scale "F" denotes enlarging the image to the preset largest size.}
    \label{tab:fd-test-scales}
    \begin{tabular}{cc|cccccccccccc} 
        \hline
        \multicolumn{1}{c|}{\multirow{2}{*}{Model}} & \multirow{2}{*}{Publication} & \multicolumn{12}{c}{test Scales (ratio)}                                                                                                                                                                                                                                                                                                                                                                                                                                                                                                                                            \\ \cline{3-14} 
        \multicolumn{1}{c|}{}                       &                              & 0.25                                                & 0.5                                                 & 0.75                                                & 1                                                   & 1.25                                                & 1.5                                                 & 1.75                                                & 2.0                                                 & 2.25                                                & S                         & E                         & F                         \\ \hline
        \multicolumn{1}{c|}{HR}                     & CVPR'17                      & \checkmark                           & \checkmark                           &                                                     & \checkmark                           &                                                     &                                                     &                                                     &                                                     &                                                     &                           &                           &                           \\
        \multicolumn{1}{c|}{S$^3$FD}                & ICCV'17                      &                                                     & \checkmark                           &                                                     & \checkmark                           &                                                     &                                                     &                                                     &                                                     &                                                     & \checkmark & \checkmark &                           \\
        \multicolumn{1}{c|}{PyramidBox}             & ECCV'18                      & \checkmark                           &                                                     & \checkmark                           & \checkmark \checkmark & \checkmark                           & \checkmark                           & \checkmark                           &                                                     &                                                     & \checkmark & \checkmark &                           \\
        \multicolumn{1}{c|}{SRN}                    & AAAI'19                      &                                                     & \checkmark                           &                                                     & \checkmark \checkmark &                                                     & \checkmark                           &                                                     &                                                     & \checkmark                           &                           &                           & \checkmark \\
        \multicolumn{1}{c|}{DSFD}                   & CVPR'19                      &                                                     & \checkmark                           &                                                     & \checkmark \checkmark & \checkmark                           &                                                     & \checkmark                           &                                                     & \checkmark                           & \checkmark & \checkmark &                           \\
        \multicolumn{1}{c|}{CSP}                    & CVPR'19                      & \checkmark \checkmark & \checkmark \checkmark & \checkmark \checkmark & \checkmark \checkmark & \checkmark \checkmark & \checkmark \checkmark & \checkmark \checkmark & \checkmark \checkmark & \checkmark \checkmark &                           &                           &                           \\ \hline
                                                    &                              & \multicolumn{12}{c}{test Scales (resize longer side)}                                                                                                                                                                                                                                                                                                                                                                                                                                                                                                                               \\ \cline{3-14} 
                                                    &                              & 100                                                 & 300                                                 & 500                                                 & 600                                                 & 700                                                 & 800                                                 & 900                                                 & 1000                                                & 1100                                                & 1200                      & 1400                      & 1600                      \\ \hline
        \multicolumn{1}{c|}{SSH}                    & ICCV'17                      &                                                     &                                                     & \checkmark                           &                                                     &                                                     & \checkmark                           &                                                     &                                                     &                                                     & \checkmark &                           & \checkmark \\
        \multicolumn{1}{c|}{SHF}                    & WACV'20                      & \checkmark                           & \checkmark                           &                                                     & \checkmark                           &                                                     &                                                     &                                                     & \checkmark                           &                                                     &                           & \checkmark &                           \\
        \multicolumn{1}{c|}{RetinaFace}             & CVPR'20                      &                                                     &                                                     & \checkmark                           &                                                     &                                                     & \checkmark                           &                                                     &                                                     & \checkmark                           &                           & \checkmark & \checkmark \\ \hline
        \end{tabular}
\end{table*}

\textbf{How do we calculate the FLOPs of different models?} We first validate whether the officially released trained models can perform as well as the authors state in their papers. It should be noted that we do not calculate the pre-processing and post-processing stages from a model's pipeline. In other words, only FLOPs of neural network layers such as convolution, activation, normalization, pooling and other layers are calculated.

Given a 4D input tensor of size $N \times C_{in} \times H_{in} \times W_{in}$ as input, a neural network layer produces a 4D output tensor of size $N\times C_{out}\times H_{out}\times W_{out}$, where $N$ is the batch size which is dismissed for simplicity in the following since it is usually set to 1 during test, $C$, $H$ and $W$ are the channels, height and width of the tensor respectively. Additionally, $K$ is introduced to represent the kernel size for layers utilizing kernels such as convolution and pooling layers. Specifically, we treat floating point operations, such as addition, subtraction, multiplication, division and exponentiation the same, which should be 1 FLOPs for simplicity. With these assumptions, we are able to derive the equations for calculating FLOPs for different layers as listed in Table~\ref{tab:flops-calc}.

We implement our FLOPs calculator based on  PyTorch regarding all the rules and equations we discussed above, which accelerates the calculation of FLOPs by dismissing any calculation related to the value of tensors, while only computing the sizes of tensors and FLOPs. This calculator can also allow us to use the code of defining models from authors with minor changes, which reduces the statistics workload. We released our source code at \url{https://github.com/fengyuentau/PyTorch-FLOPs}.

\begin{figure*}[htbp]
    \begin{center}
        \includegraphics[width=1.0\linewidth]{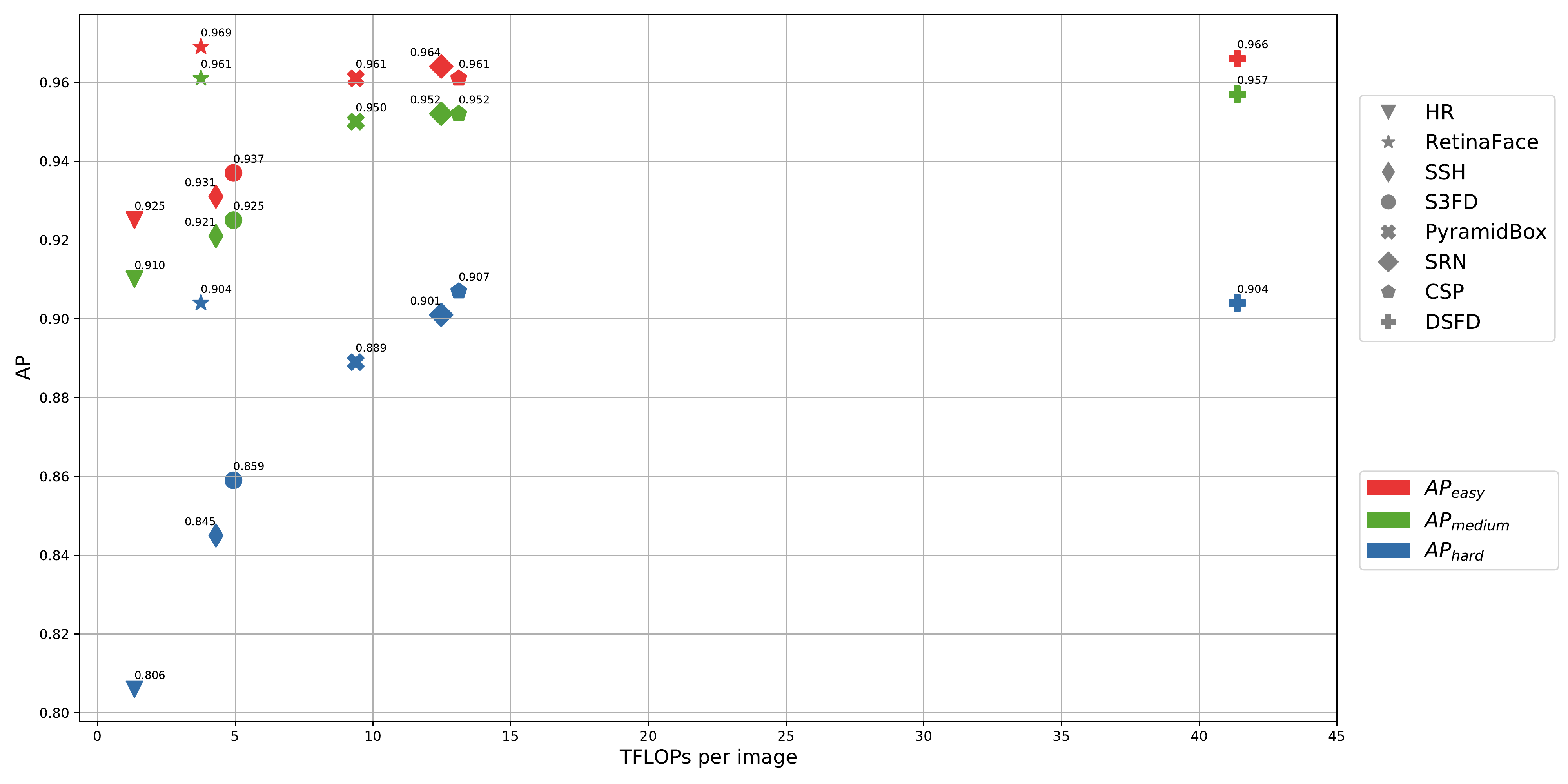}
    \end{center}
    \caption{
        The FLOPs vs. multi-scale AP of WIDER Face validation set. 7 models from the WIDER Face result page are listed, which are 
        HR~\cite{fd-hr}, SSH~\cite{fd-ssh}, S$^3$FD~\cite{fd-s3fd}, PyramidBox~\cite{fd-pyramidbox}, SRN~\cite{fd-srn}, DSFD~\cite{fd-dsfd}, CSP~\cite{fd-csp}. (\textit{The TFLOPs for some speed-focusing face detectors are listed in Table~\ref{tab:speedy-detectors-comparison} because the TFLOPs are in a much smaller scale and cannot fit in this figure.})}
    \label{fig:val_ap_comp}
\end{figure*}

\begin{figure*}[htbp]
    \begin{center}
        \includegraphics[width=1.0\linewidth]{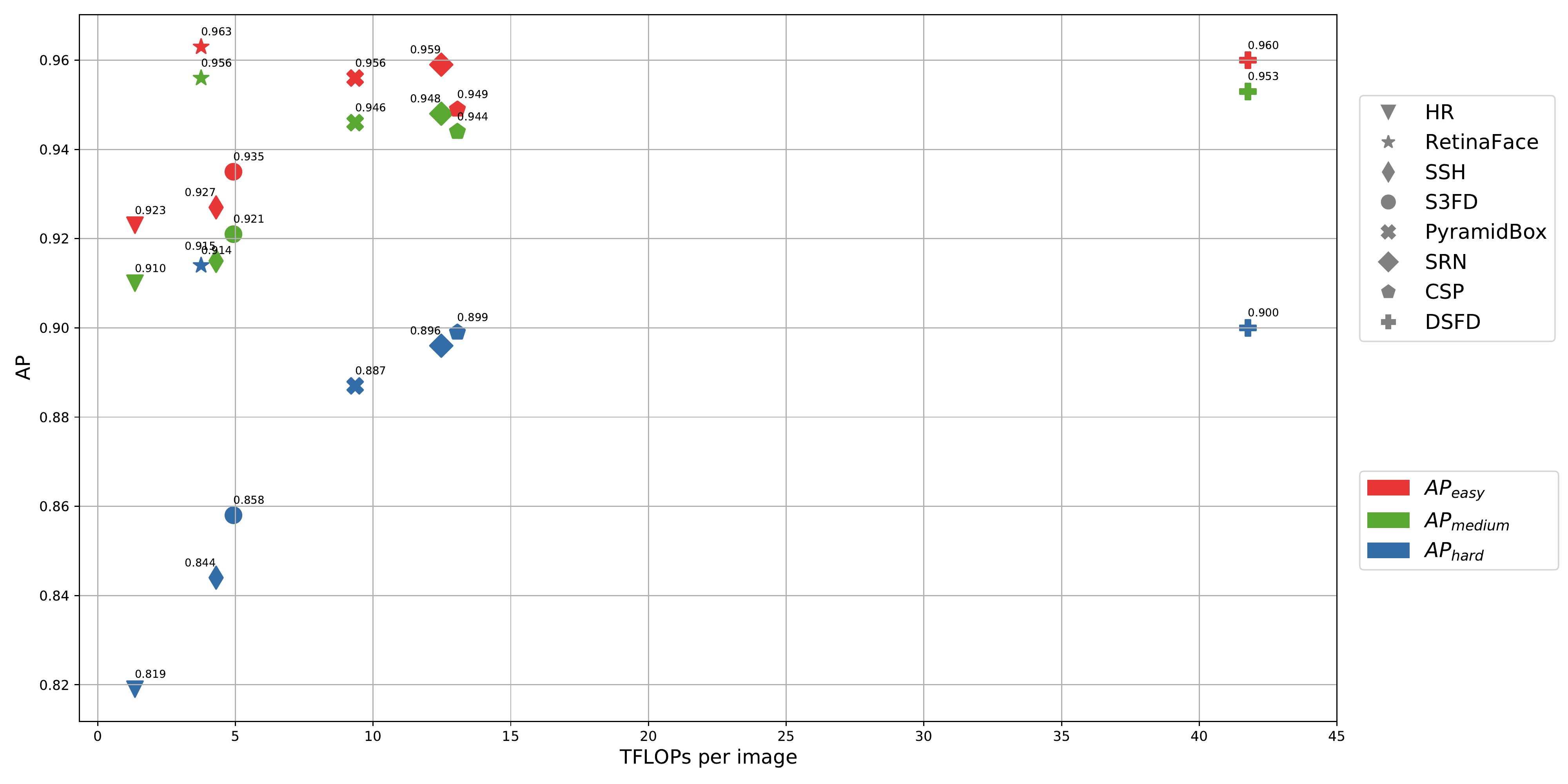}
    \end{center}
    \caption{
        The FLOPs vs. multi-scale test AP of WIDER Face test set. 7 models from the WIDER Face result page are listed, which are 
        HR~\cite{fd-hr}, SSH~\cite{fd-ssh}, S$^3$FD~\cite{fd-s3fd}, PyramidBox~\cite{fd-pyramidbox}, SRN~\cite{fd-srn}, DSFD~\cite{fd-dsfd}, CSP~\cite{fd-csp}.}
    \label{fig:test_ap_comp}
\end{figure*}

\subsection{FLOPs vs. AP in Multi-Scale Test}


The multi-scale test metric is to test a model with a set derived from an image at original and different scales (with aspect ratio fixed). The detection results of different scales are then merged and applied with the non-maximum suppression (NMS), so as to suppress the overlapped bounding boxes and reduce false positives. Based on the training data and scheme, a \textit{comfort zone} of a model is determined, which is a range of scales of faces that can be detected. The multi-scale test metric can improve a model's AP by re-scaling out-of-zone faces back in the comfort zone. However, since we cannot determine which of the faces in the test set are out-of-zone, we have to apply re-scaling to every image in the set. It leads to the multiplied increase in FLOPs per image.

\begin{table*}[htbp]
    \centering
    \caption{How different scales impact the AP of PyramidBox~\cite{fd-pyramidbox}. We use Scale = {1} as the baseline, and then try adding different scales one by one to test how AP is impacted by different scales.}
    \label{tab:pyramidbox-1}
    \begin{tabular}{llllll|lll|l}
    \multicolumn{6}{c|}{Test Scales}              & \multirow{2}{*}{$AP_{easy}$} & \multirow{2}{*}{$AP_{medium}$} & \multirow{2}{*}{$AP_{hard}$} & \multirow{2}{*}{TFLOPs} \\
    0.25 & 0.75 & 1    & 1.25 & 1.5  & 1.75 &                           &                             &                           &                         \\ \hline
         &      & \checkmark &      &      &      & 0.947                     & 0.936                       & 0.875                     & 1.37                    \\ \hline
    \checkmark &      & \checkmark &      &      &      & 0.954(+0.007)             & 0.939(+0.003)               & 0.872(-0.003)             & 1.45(+0.008)            \\
         & \checkmark & \checkmark &      &      &      & 0.952(+0.005)             & 0.940(+0.004)               & 0.874(-0.001)             & 2.14(+0.77)             \\
         &      & \checkmark & \checkmark &      &      & 0.948(+0.001)             & 0.938(+0.002)               & 0.884(+0.009)             & 2.72(+1.35)             \\
         &      & \checkmark &      & \checkmark &      & 0.947(+0.000)             & 0.937(+0.001)               & 0.881(+0.006)             & 2.46(+1.09)             \\
         &      & \checkmark &      &      & \checkmark & 0.946(-0.001)             & 0.936(+0.000)               & 0.874(-0.001)             & 1.63(+0.26)            
    \end{tabular}
\end{table*}

\begin{table*}[htbp]
    \centering
    \caption{How much will AP and FLOPs decrease if a scale is removed? The detector PyramidBox is employed.}
    \label{tab:pyramidbox-2}
    \begin{tabular}{llllll|lll|l}
    \multicolumn{6}{c|}{Test Scales}              & \multirow{2}{*}{$AP_{easy}$}                                & \multirow{2}{*}{$AP_{medium}$}        & \multirow{2}{*}{$AP_{hard}$} & \multirow{2}{*}{TFLOPs} \\
    0.25 & 0.75 & 1    & 1.25 & 1.5  & 1.75 &                                                          &                                    &                           &                         \\ \hline
    \checkmark & \checkmark & \checkmark & \checkmark & \checkmark & \checkmark & 0.957 & \multicolumn{1}{l|}{0.945}         & 0.886                     & 4.94                    \\ \hline
         & \checkmark & \checkmark & \checkmark & \checkmark & \checkmark & 0.949(-0.008) & \multicolumn{1}{l|}{0.940(-0.005)} & 0.884(-0.002) & 4.85(-0.009)            \\
    \checkmark &      & \checkmark & \checkmark & \checkmark & \checkmark & 0.954(-0.003) & \multicolumn{1}{l|}{0.942(-0.003)} & 0.885(-0.001) & 4.16(-0.780)            \\
    \checkmark & \checkmark &      & \checkmark & \checkmark & \checkmark & 0.955(-0.002) & \multicolumn{1}{l|}{0.940(-0.005)} & 0.850(-0.013) & 3.58(-1.360)             \\
    \checkmark & \checkmark & \checkmark &      & \checkmark & \checkmark & 0.957(+0.000) & \multicolumn{1}{l|}{0.944(-0.001)} & 0.880(-0.006) & 3.58(-1.360)             \\
    \checkmark & \checkmark & \checkmark & \checkmark &      & \checkmark & 0.958(+0.001) & \multicolumn{1}{l|}{0.945(+0.000)} & 0.884(-0.002) & 3.84(-1.100)              \\
    \checkmark & \checkmark & \checkmark & \checkmark & \checkmark &      & 0.957(+0.000) & \multicolumn{1}{l|}{0.945(+0.000)} & 0.886(+0.000) & 4.67(-0.270)            
    \end{tabular}
\end{table*}

Fig.~\ref{fig:val_ap_comp} and Fig.~\ref{fig:test_ap_comp} show the multi-scale test AP and FLOPs of different models on the validation and test sets of the WIDER Face dataset, respectively. 
We can find a clear trend in the two figures. The FLOPs are increasing and the AP is improving in the sequence of methods HR~\cite{fd-hr}, SSH~\cite{fd-ssh}, S$^3$FD~\cite{fd-s3fd}, PyramidBox~\cite{fd-pyramidbox}, SRN~\cite{fd-srn} and CSP~\cite{fd-csp}. There are two methods do not follow the trend. The first one is DSFD~\cite{fd-dsfd} which has more than 3 times of FLOPs than  SRN and CSP, but the AP is similar with those of SRN and CSP. It means DSFD has unreasonable high computational cost. Then second detector is RetinaFace~\cite{retinaface} which gained the best AP but the computational cost is much lower than most other methods.

The two figures (Fig.~\ref{fig:val_ap_comp} and Fig.~\ref{fig:test_ap_comp}) give us a clear view of different face detection models and can guide us understand different models deeper.


\subsection{FLOPs vs. AP in Single-Scale Test}

FLOPs can sharply increase in two ways: fundamentally increasing through introducing more complex modules to the network, and through multi-scale testing. As Table~\ref{tab:fd-test-scales} shows, these models are all tested on various scales. However, why models are tested on these various scales is seldom discussed. How much contribution on AP can one scale bring? Are any scales not necessary?

\textbf{Single-scale test on a single model}. Table~\ref{tab:pyramidbox-1} shows the AP contribution of different scales. The easy subset in WIDER Face~\cite{fd-widerface} contains a large margin of faces of regular size and some large faces, as a result of which shrinking images can help improve the AP. We can observe that $AP_{hard}$ gains the most from scales ${1, 1.25}$ and ${1, 1.5}$ , but not for scale ${1, 1.75}$. Together with FLOPs, we can also observe an increase to the peak at scale ${1, 1.25}$ and then a sharp drop for larger scales. The reason is that a threshold for the largest size of images is set to avoid exceeding the GPU memory. This means that not all 1.75x resized images were sent to a detector in the experiments.

Table~\ref{tab:pyramidbox-2} shows how much the AP and FLOPs will decrease if a model tested without a scale. As the missing scale becomes larger, the decrease of $AP_{easy}$ decreases. However, this pattern does not apply to $AP_{medium}$ and $AP_{hard}$. The reason is that the enlarged images will be skipped if their size goes beyond the preset limit, so as to avoid exceeding GPU memory. The larger the scale is, the fewer images will be re-scaled and tested. The drop of FLOPs greatly decreases on scale 1.75. This is because the PyramidBox pretrained model is mainly trained on scale 1.

The two tables~\ref{tab:pyramidbox-1} and ~\ref{tab:pyramidbox-2} imply that $AP_{easy}$ is the most sensitive to scales 0.25, $AP_{medium}$ is the most sensitive to scale 0.25 and 1, and $AP_{hard}$ is the most sensitive to scale 1. Note that this is highly related to the training scale. If the model is trained differently, the conclusion may change accordingly. 

\textbf{Single-scale test on multiple models}.

Table~\ref{tab:multi-models-single-scale} shows the AP and FLOPs of different models on scale 1. The large overall leap is brought by PyramidBox~\cite{fd-pyramidbox}, which mainly introduces the FPN~\cite{fpn} module to fuse features from two adjacent scales and the context enhancing module from SSH~\cite{fd-ssh}. The computational cost of PyramidBox is ~2X compared with SSH but less than 1/2 of DSFD. However, the AP achieved by PyramidBox and DSFD are comparable.


\begin{table}[htp]
    \centering
    \caption{AP and FLOPs of different models on scale 1.}
    \label{tab:multi-models-single-scale}
    \begin{tabular}{c|ccc|c}
    Model      & $AP_{easy}$ & $AP_{medium}$ & $AP_{hard}$ & TFLOPs \\ \hline
    RetinaFace & 0.952    & 0.942      & 0.776    & 0.198   \\
    S$3$FD     & 0.924    & 0.906      & 0.816    & 0.571   \\
    CSP        & 0.948    & 0.942      & 0.774    & 0.571   \\
    SSH        & 0.925    & 0.909      & 0.731    & 0.587   \\
    PyramidBox & 0.947    & 0.936      & 0.875    & 1.387   \\
    DSFD       & 0.949    & 0.936      & 0.845    & 1.532   
    \end{tabular}
\end{table}

If some benchmarks can evaluate FLOPs or some other similar efficiency measurements, different face detectors can compare more fairly. It will also promote face detection research to a better stage.

\subsection{FLOPs vs Latency}
To compare the two measurements, we convert existing models to the Open Neural Network Exchange (ONNX) format and run them using the ONNXRUNTIME\footnote{\url{https://github.com/microsoft/onnxruntime}} in this comparison for fair comparison. Note that due to the different supports to ONNX converting of different DL frameworks, we managed to convert RetinaFace~\cite{retinaface}, SRN~\cite{fd-srn}, DSFD~\cite{fd-dsfd} and CSP~\cite{fd-csp} to ONNX format. The results are in Table~\ref{tab:flops-time}. These models are evaluated using an NVIDIA QUADRO RTX 6000 with CUDA 10.2, and an INTEL Xeon Gold 6132 CPU @ 2.60 GHz. The powerful GPU contains 4,609 CUDA parallel-processing cores and 24GB memory.

We can observe that both FLOPs and forward latency increase from RetinaFace~\cite{retinaface} to DSFD~\cite{fd-dsfd}. Note that although the average FLOPs of RetinaFace are just one-fifth of SRN's, the forward latency of RetinaFace is almost near half of SRN's, implying that FLOPs are not linearly correlated to latency due to the differences in implementation, hardware settings, memory efficiency and so on. The reason why the post-processing latency of DSFD and CSP sharply increase is that they do not use GPU-accelerated NMS as others do.

\begin{table}[htbp]
    \centering
    \caption{
        State-of-the-art open-source models tested with a 720P image containing several faces at scale=1.0 only. We average the FLOPs (AVG TFLOPs) and latency (AVG Latency) by running the test for each model 100 times. Note that 'Post-Proc' denotes post-processing stages, such as decoding from anchors, NMS and so on. For this stage, we adopt the original processing code of each model.}
    \label{tab:flops-time}
    \begin{tabular}{|c|c|c|c|c|}
    \hline
    \multirow{2}{*}{Model} & \multirow{2}{*}{\begin{tabular}[c]{@{}c@{}}AVG\\ TFLOPs\end{tabular}} & \multicolumn{3}{c|}{AVG Latency (ms)}                                                                                           \\ \cline{3-5} 
                           &                                                                       & \begin{tabular}[c]{@{}c@{}}Forward\\ (GPU)\end{tabular} & \begin{tabular}[c]{@{}c@{}}Forward\\ (CPU)\end{tabular} & Post-Proc   \\ \hline
    RetinaFace  & 0.201   & 131.60   & 809.24   & 8.74 (GPU)  \\ \hline
    CSP         & 0.579   & 154.55   & 1955.20  & 27.74 (CPU) \\ \hline
    SRN         & 1.138   & 204.77   & 2933.16  & 8.71 (GPU)  \\ \hline
    DSFD        & 1.559   & 219.63   & 3671.46  & 76.32 (CPU) \\ \hline
    \end{tabular}
\end{table}


\section{Speed-Focusing Face Detectors} \label{review_speed}

\begin{table*}[htbp]
    \centering
    \caption{
        Popular and active open-source face detectors at Github. Note that 'AVG GFLOPs' are computed on WIDER Face validation set in single-scale test where only scale=1.0. Also note that latency is measured on CPU.
    }
    \label{tab:speedy-detectors-comparison}
    \begin{tabular}{|c|c|c|c|c|c|c|c|c|}
    \hline
    \multirow{2}{*}{Model} & \multirow{2}{*}{\begin{tabular}[c]{@{}c@{}}\#CONV\\ Layers\end{tabular}} & \multirow{2}{*}{\begin{tabular}[c]{@{}c@{}}\#Params\\ ($\times10^6$)\end{tabular}} & \multirow{2}{*}{\begin{tabular}[c]{@{}c@{}}AVG\\ GFLOPs\end{tabular}} & \multicolumn{3}{c|}{WIDER Face Val Set}   & \multicolumn{2}{c|}{Latency (ms)} \\ \cline{5-9} & & & & $AP_{easy}$ & $AP_{medium}$ & $AP_{hard}$ & Forward & Post-Proc \\ \hline
    FaceBoxes~\cite{FaceBoxes}              & 33 & 1.013 & 1.541  & 0.845 & 0.777 & 0.404 & 16.52  & 7.16 \\ \hline
    ULFG-slim-320~\cite{ULFG}               & \multirow{2}{*}{42} & \multirow{2}{*}{0.390} & \multirow{2}{*}{2.000} & 0.652 & 0.646 & 0.520 & \multirow{2}{*}{19.03} & \multirow{2}{*}{2.37} \\ \cline{1-1} \cline{5-7}
    ULFG-slim-640~\cite{ULFG}               & & & & 0.810 & 0.794 & 0.630 & & \\ \cline{1-9}
    ULFG-RFB-320~\cite{ULFG}                & \multirow{2}{*}{52} & \multirow{2}{*}{0.401} & \multirow{2}{*}{2.426} & 0.683 & 0.678 & 0.571 & \multirow{2}{*}{21.27} & \multirow{2}{*}{1.90} \\ \cline{1-1} \cline{5-7}
    ULFG-RFB-640~\cite{ULFG}                & & & & 0.816 & 0.802 & 0.663 & & \\ \hline
    YuFaceDetectNet~\cite{YuFaceDetectNet}  & 43 & 0.085 & 2.549  & 0.856 & 0.842 & 0.727 & 23.47  & 32.81 \\ \hline
    LFFD-v2~\cite{LFFD}                     & 45 & 1.520 & 37.805 & 0.875 & 0.863 & 0.752 & 178.47 & 6.70 \\ \hline
    LFFD-v1~\cite{LFFD}                     & 65 & 2.282 & 55.555 & 0.910 & 0.880 & 0.778 & 229.35 & 10.08 \\ \hline
    \end{tabular}
\end{table*}

For the face detectors introduced in the previous sections, the main target is to reach a better AP. Their computational costs are heavy and normally in magnitude of TFLOPs. It is unrealistic to deploy those heavy models to a face-related system. There are some other open source face detectors whose target is to make face detection run in real time for practical applications. Their computational costs are in the magnitude of those of GFLOPs or 10 GFLOPs and are much less than the previous costs. Here we group them as speed-focusing face detectors. We collect the most-popular ones from \url{github.com}, and review them in terms of network architectures, AP, FLOPs and efficiency.

\textbf{FaceBoxes}~\cite{FaceBoxes} is one of the first one-stage deep learning-based models to achieve real-time face detection. FaceBoxes rapidly downsamples feature maps to a stride 32 with two convolution layers with large kernels. Inception blocks~\cite{Inception-V1} are introduced to enhanced feature maps at stride of 32. Following the multi-scale mechanism from SSD~\cite{ssd}, FaceBoes detects on layers \texttt{inception3}, \texttt{conv3\_2} and \texttt{conv4\_2} for faces at different scales, resulting in an AP of 0.960 on FDDB~\cite{fd-fddb} and 20 FPS on an INTEL E5-2660v3 CPU at 2.60 GHz.

\textbf{YuFaceDetectNet}~\cite{YuFaceDetectNet} adopts a light MobileNet~\cite{MobileNet-v1} as the backbone. Compared to FaceBoxes, YuFaceDetectNet has more convolution layers on each stride to have fine-grained features, and detects on the extra layer of stride 16, which improves the recall of small faces. The evaluation results of the model on the WIDER Face~\cite{fd-widerface} validation set are 0.856 (Easy), 0.842 (Medium) and 0.727 (Hard). The main and well-known repository, libfacedetection~\cite{libfacedetection}, takes YuFaceDetectNet as the detection model and offers pure C++ implementation without dependence on DL frameworks, resulting from 77.34 FPS for $640 \times 480$ images to 2,027.74 FPS for $128 \times 96$ images on an INTEL i7-1065G7 CPU at 1.3 GHz.

\textbf{LFFD}~\cite{LFFD} introduces residual blocks for feature extraction, and proposes receptive fields as the natural anchors. Its faster version LFFD-v2 managed to achieve 0.875 (Easy), 0.863 (Medium) and 0.754 (Hard) on the WIDER Face validation set, while running at 472 FPS using CUDA 10.0 and an NVIDIA RTX 2080Ti GPU. \textbf{ULFG}~\cite{ULFG} adds even more convolution layers on each stride, taking the advantage of depth-wise convolution, which is friendly to edge devices in terms of FLOPs and forward latency. As reported, the slim version of ULFG has an AP of 0.770 (Easy), 0.671 (Medium) and 0.395 (Hard) on the WIDER Face validation set, and can run at 105 FPS with an input resolution of $320 \times 240$ on an ARM A72 at 1.5 GHz.

These light-weight models are developed using various frameworks and tested on different hardware. For fair comparison, we export these models from their original frameworks to ONNX and test using ONNXRUNTIME on a INTEL i7-5930K CPU at 3.50GHz. Results are shown in Table~\ref{tab:speedy-detectors-comparison}. We can observe that more \texttt{CONV} layers do not lead to more parameters (FacesBoxes and ULFG series) and more FLOPs (YuFaceDetectNet and ULFG series). This is mainly because of the extensive usage of depth-wise convolution in ULFG. Additionally, note that more FLOPs do not lead to more forward latency due to depth-wise convolution. The post-processing latency across different face detectors seems inconsistent with the forward latency, and we verified that this is caused by different numbers of bounding boxes sent to NMS and the different implementations of NMS (Python-based or Cython-based).

\section{Conclusions and Discussions} \label{conclusion}

Face detection is one of the most important and popular topics yet still challenging in computer vision. Deep learning has brought remarkable breakthroughs for face detectors. Face detection is more robust and accurate even in unconstrained real-world environments. In this paper, recent deep learning-based face detectors and benchmarks are introduced. From the evaluations of accuracy and efficiency on different deep  face detectors, we can find that we can reach a very high accuracy if we do not consider the computational cost. However, there should be a simple and beautiful solution for face detection since it is simpler than generic object detection. The research on face detection can focus on the topics introduced in the following topics in the future.

\textbf{Superfast Face Detection}. There is no definition for superfast face detection. Ideally, superfast face detector should be able to run in real time on low-cost edge devices even when the input image is 1080P. Empirically speaking, we would like to expect it to be less than 100M FLOPs with a 1080P image as input. For real-world applications, efficiency is one of the key issues. Efficient face detectors can help to save both energy, the cost of hardware and improve the responsiveness for edge devices, such as CCTV cameras and mobile phones.

\textbf{Detecting Faces in the Long-tailed Distribution}. Face samples can be regarded as a long-tailed distribution. Most face detectors are trained for the dominant part of the distribution. We have already had enough samples for faces with variances in illumination, pose, scale,  occlusion, blur, distortion in the WIDER Face dataset. But what about other faces like the old and damaged ones? As people getting old, there are many wrinkles on their faces; and people who suffer from illnesses or accidents may have damaged faces, such as burn scars on the faces. Face detection is not only a technical problem but also a humanitarian problem, meaning that this technology should serve all the people, not only the dominant part of the population. Ideally, face detectors should be able to detect all kinds of faces. However, in most face datasets and benchmarks, most faces are from young people.

The final goal of face detection is to detect faces with very high accuracy and high efficiency. Therefore, the algorithms can be deployed to many kinds of edge devices and centralized servers to improve the perception capability of computers; currently, there still is a considerable gap. Face detectors can achieve good accuracy but still require considerable computations. Improving the efficiency should be the next step.

\ifCLASSOPTIONcaptionsoff
  \newpage
\fi

\bibliographystyle{IEEEtran}
\bibliography{reference}

\begin{IEEEbiography}[{\includegraphics[height=30mm,clip,keepaspectratio]{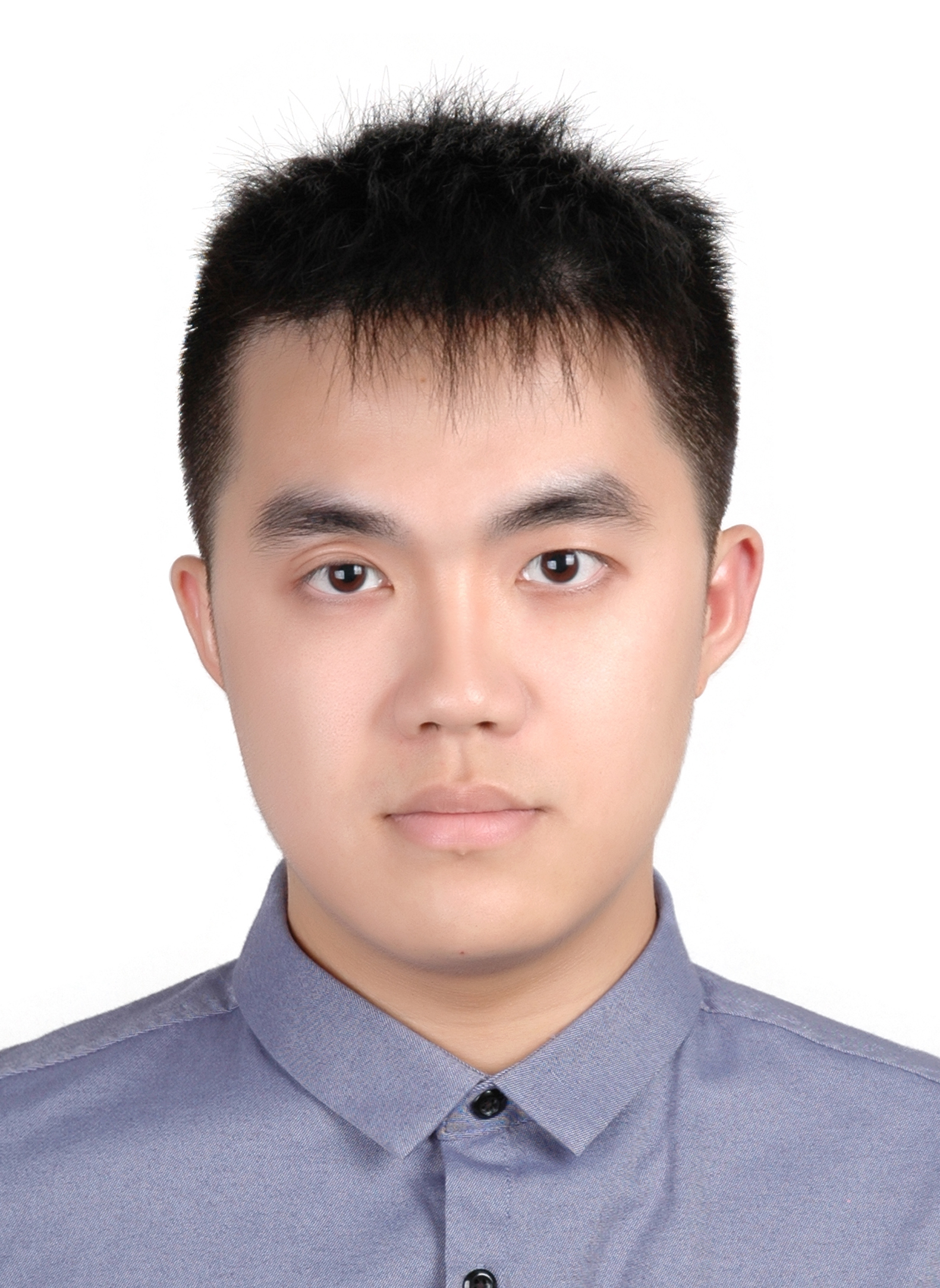}}]{Yuantao Feng} is currently a Research Assistant in the Department of Computer Science and Engineering, Southern University of Science and Technology, China. He received his B.E. degree and M.E. in computer science and technology from the College of Computer and Software Engineering, Shenzhen University in 2018 and 2021 respectively. His research interests include object detection and computer vision.
\end{IEEEbiography}

\begin{IEEEbiography}[{\includegraphics[height=30mm,clip,keepaspectratio]{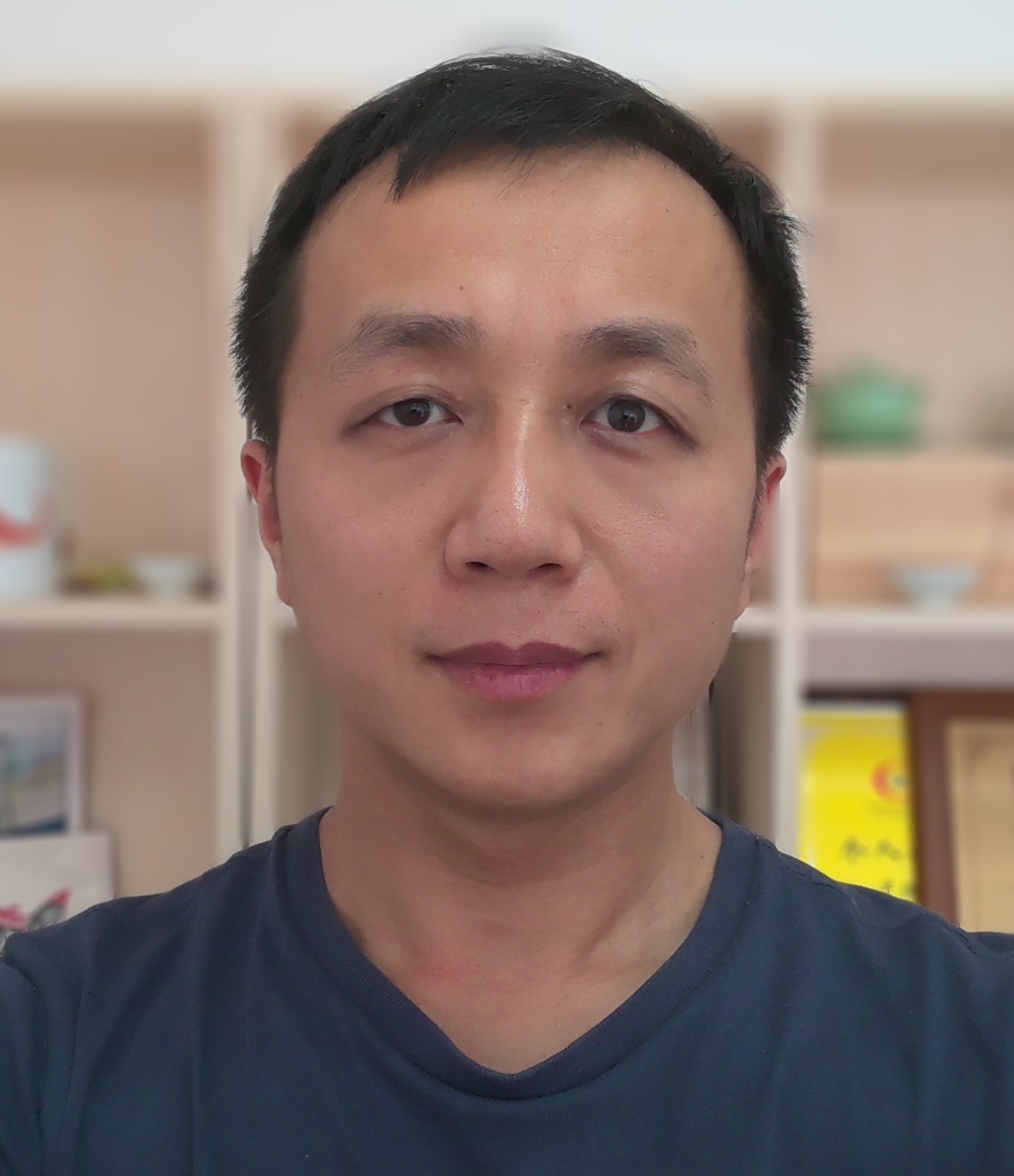}}]{Shiqi Yu} is currently an Associate Professor in the Department of Computer Science and Engineering, Southern University of Science and Technology, China. He received his B.E. degree in computer science and engineering from the Chu Kochen Honors College, Zhejiang University in 2002, and Ph.D. degree in pattern recognition and intelligent systems from the Institute of Automation, Chinese Academy of Sciences in 2007. He worked as an Assistant Professor and an Associate Professor in Shenzhen Institutes of Advanced Technology, Chinese Academy of Sciences from 2007 to 2010, and as an associate professor in Shenzhen University from 2010 to 2019. His research interests include gait recognition, face detection and computer vision.
\end{IEEEbiography}

\begin{IEEEbiography}[{\includegraphics[height=30mm,clip,keepaspectratio]{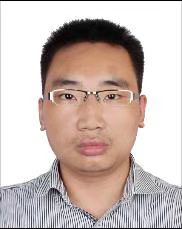}}]{Hanyang Peng} received a B.S. degree in measurement and control technology from the Northeast University of China, Shenyang, China, in 2008, an M.E. degree in detection technology and automatic equipment from the Tianjin University of China, Tianjin, China, in 2010, and a Ph.D. degree in pattern recognition and intelligence systems from the Institute of Automation, Chinese Academy of Sciences, Beijing, China, in 2017. He is currently with Southern University of Science and Technology, Shenzhen, China. His current research interests include computer vision, machine learning, deep learning and optimization.
\end{IEEEbiography}

\begin{IEEEbiography}[{\includegraphics[height=30mm,clip,keepaspectratio]{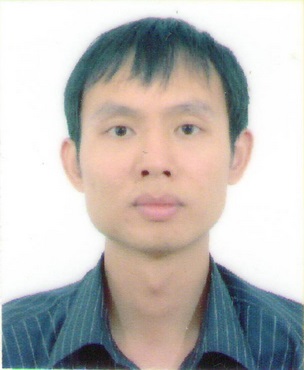}}]{Yan-Ran Li} is currently an Associate Professor in the College of Computer Science and Software Engineering, Shenzhen University, China. He received a Ph.D. degree in communications and information systems from the Electronic Department, School of Information Science and Technology, Sun Yat-Sen (Zhongshan) University, China, in 2009. He was a Visiting Scholar with the National University of Singapore, Singapore, in 2008, and  the Chinese University of Hong Kong, Hong Kong, in 2011 and 2012, and Syracuse University, USA, in 2014. His current research interests include image processing and medical data processing.
\end{IEEEbiography}

\begin{IEEEbiography}[{\includegraphics[width=1in,height=1.25in,clip,keepaspectratio]{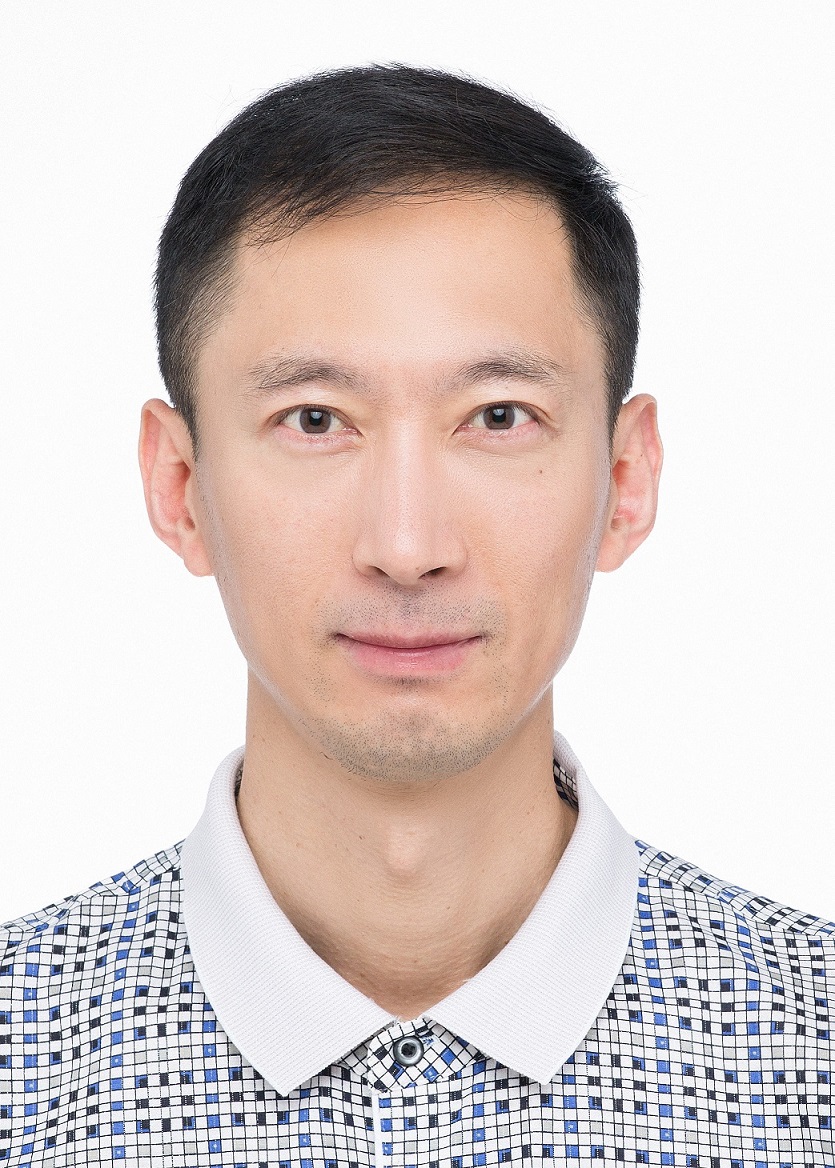}}]{Jianguo Zhang} is currently a Professor in the Department of Computer Science and Engineering, Southern University of Science and Technology. Previously, he was a Reader in Computing, School of Science and Engineering, University of Dundee, UK. He received a PhD in the National Lab of Pattern Recognition, Institute of Automation, Chinese Academy of Sciences, Beijing, China, 2002. His research interests include object recognition, medical image analysis, machine learning and computer vision. He is a senior member of the IEEE and serves as an Associate Editor of IEEE Trans on Multimedia.
\end{IEEEbiography}

\end{document}